\pdfoutput=1

\documentclass{article}
\usepackage{arxiv} 
\usepackage{url} 
\usepackage{eucal}
\usepackage[T1]{fontenc}
\usepackage{amsfonts}
\usepackage{booktabs}  
\usepackage{microtype}      
\usepackage{lipsum}
\usepackage{graphicx}
\usepackage{amsmath}
\usepackage[normalem]{ulem}
\usepackage{xcolor}
\usepackage{xspace}
\usepackage{algorithm2e}
\usepackage{varwidth}
\usepackage{algpseudocode}
\usepackage{multirow}
\usepackage{colortbl}
\usepackage{makecell}
\usepackage{subcaption}
\usepackage{tabularx}
\usepackage[size=tiny]{todonotes}
\usepackage{hyperref}
\usepackage{natbib}

\newcommand{\se}[1]{\textcolor{black}{#1}}
\newcommand{\zr}[1]{\textcolor{black}{#1}}
\newcommand{\vect}[1]{\boldsymbol{#1}}

\newcommand{\bygpt}{\textsc{ByGPT5}\xspace}
\newcommand{\quatrain}{\textsc{QuaTrain}\xspace}
\newcommand{\llmagent}{LLM-based agents\xspace}
\newcommand{\ingroup}{`in-group'\xspace}
\newcommand{\outgroup}{`out-group'\xspace}
\newcommand{\lce}{$\mathcal{L}_{\text{CE}}$\xspace}

\newcommand{\lcl}{$\mathcal{L}_{\text{CL}}$\xspace}
\newcommand{\trainable}{\textsc{training-based}\xspace}
\newcommand{\untrainable}{\textsc{prompting-based}\xspace}
\newcommand{\ce}{\textsc{Cross-Entropy}\xspace}
\newcommand{\dex}{\textsc{DExpert}\xspace}
\newcommand{\fgenerate}{F_{\text{generate}}\xspace}
\newcommand{\fupdate}{F_{\text{update}}\xspace}
\newcommand{\aij}{\mathcal{A}_i\xspace}
\newcommand{\aijout}{\mathcal{A}^{-}\xspace}
\newcommand{\aijin}{\mathcal{A}^{+}\xspace}
\newcommand{\pa}{P_{{\mathcal{A}_{i}}}\xspace}
\newcommand{\pan}{P_{*^-}\xspace}
\newcommand{\pap}{P_{*^+}\xspace}
\newcommand{\pup}{\textsc{Update}\xspace}
\newcommand{\pgen}{\textsc{Generate}\xspace}

\newcommand{\contra}{\textsc{Contrastive}\xspace}
\newcommand{\sbert}{\textsc{SBERT}\xspace}
\newcommand{\pchain}{chain-prompting\xspace}
\newcommand{\pjoint}{joint-prompting\xspace}

\makeatletter
\newcommand{\algrule}[1][.2pt]{\par\vskip.5\baselineskip\hrule height #1\par\vskip.5\baselineskip}
\makeatother

\title{LLM-based multi-agent poetry generation in non-cooperative environments}

\author{
 Ran Zhang \\
 School of Business Informatics and Mathematics\\
 University of Mannheim \\
 B6 26, 68159 Mannheim Germany\\
 \texttt{ran.zhang@uni-mannheim.de}\\
   \And
Steffen Eger \\
  School of Business Informatics and Mathematics\\
 University of Mannheim \\
 B6 26, 68159 Mannheim Germany\\
 \texttt{steffen.eger@uni-mannheim.de}\\
}

\begin{document}
\maketitle
\begin{abstract}
Despite substantial progress of large language models (LLMs) for automatic poetry generation, the generated poetry lacks diversity while the training process differs greatly from human learning. Under the rationale that the learning process of the poetry generation systems should be more human-like and their output more diverse and novel, we introduce a framework based on social learning where we emphasize non-cooperative interactions besides cooperative interactions to encourage diversity. Our experiments are the first attempt at LLM-based multi-agent systems in non-cooperative environments for poetry generation employing both \trainable agents (GPT-2) and \untrainable agents (GPT-3 and GPT-4). Our evaluation based on 96k generated poems shows that our framework benefits the poetry generation process for \trainable agents resulting in 1) a 3.0-3.7 percentage point  (pp) increase in diversity and a 5.6-11.3 pp increase in novelty according to distinct and novel n-grams. The generated poetry from \trainable agents also exhibits group divergence in terms of lexicons, styles and semantics.
\untrainable agents in our framework also benefit from non-cooperative environments and a more diverse ensemble of
models with non-homogeneous agents has the potential to further enhance diversity, with an increase of 7.0-17.5 pp according to our experiments. However, \untrainable agents show a decrease in lexical diversity over time and do not exhibit the group-based divergence intended in the social network.
Our paper argues for a paradigm shift in creative tasks such as automatic poetry generation to include social learning processes (via LLM-based agent modeling) similar to human interaction. 
\end{abstract}


\section{Introduction}
\label{sec:introduction}

Autonomous agents driven by large language models (LLMs) have made substantial progress in various domains including complex task-solving \cite{li2024camel}, reasoning \cite{lin2024swiftsage, du2023improving}, and simulation \cite{wang2023voyager}. Studies have shown that interactive communication via mono- or multi-agent systems can yield emergent behaviors \cite{park2023generative}, enhanced task performance \cite{zhuge2023mindstorms}, better evaluation \cite{chan2023chateval}, and assistance in open-end generation tasks \cite{zhu2023calypso}, to name a few. Despite these advancements, the exploration of creative tasks such as poetry generation utilizing LLM-based agents is still limited \cite{chakrabarty2023creative}. This paper presents the first experiment on LLM-based multi-agent poetry generation.\footnote{The code, data and generated outputs are publicly available at \url{https://github.com/zhangr2021/Multiagent_poetry.git}} We introduce a framework that emphasizes non-cooperative environments to enhance diversity and novelty in generated poetry both in aggregative mean over time and dynamically across iterations. \newline

\begin{figure}
    \centering    \includegraphics[width=0.95\textwidth]{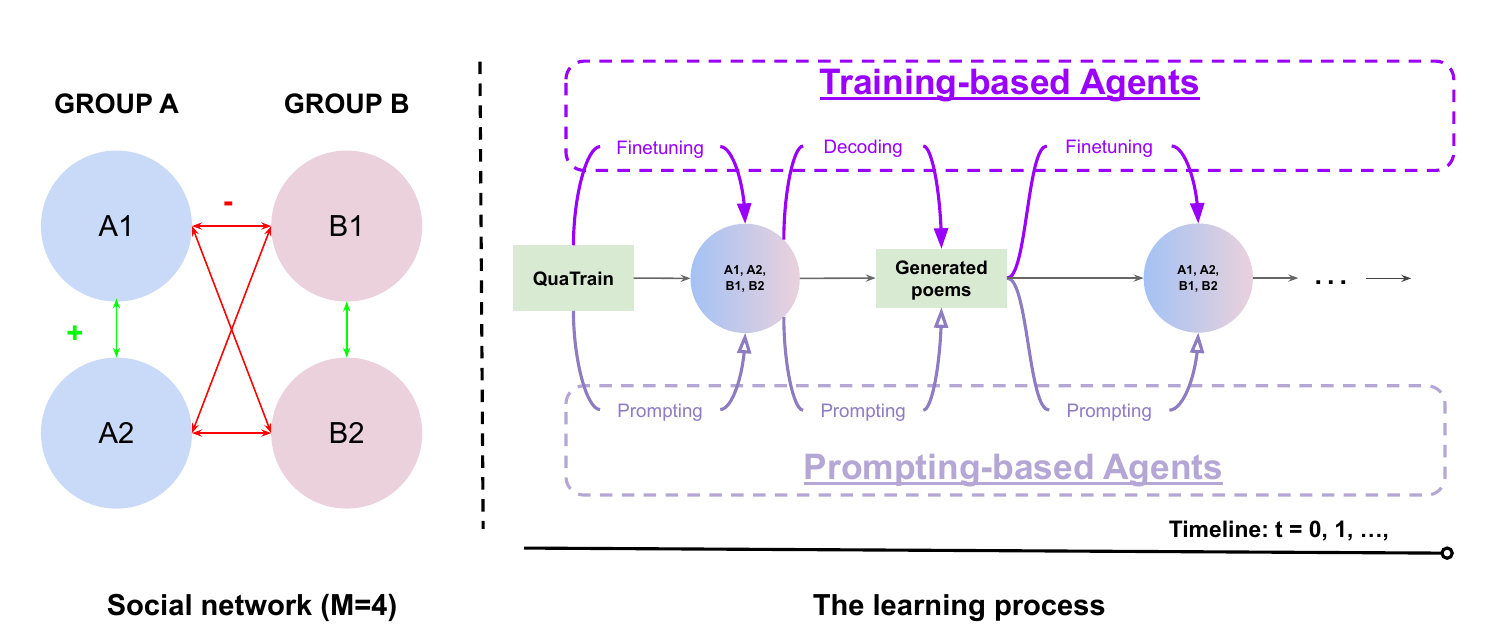}
    \caption{Illustration of the predefined social network ($M=4$) and the high-level description of the learning process for \trainable agents (GPT-2) and \untrainable agents (GPT-3.5 and GPT-4). The \textcolor{green}{green} and \textcolor{red}{red} lines in the social network indicate cooperative (\textcolor{green}{+}) and non-cooperative (\textcolor{red}{-}) interaction between agents respectively.}
    \label{fig:network}
\end{figure}

\noindent\textbf{Why poetry generation?} \\

{\centering ``Poetry is the rhythmical creation of beauty in words.'' \\
\hspace{8cm}\textit{-- Edgar Allan Poe}} \newline

\noindent Despite advancements in LLMs, generating poetry remains a difficult task due to the complex interplay of style, meaning, and human emotion \cite{chakrabarty2021don, mahbub2023unveiling}. Models need to excel not only in linguistic competence such as understanding semantics and grammar but also in capturing stylistic elements such as rhyme, meter, imagery, and artistic flair to produce human-like poetry \cite{zhipeng-etal-2019-jiuge, belouadi-eger-2023-bygpt5, ma-etal-2023-yu}. Current challenges in poetry generation include: 1) limitations of finetuned models or pipelines that are tailored for specific styles or topics \cite{lau-etal-2018-deep, van-de-cruys-2020-automatic, tian2022zero}; 2) LLMs struggling to create diverse and aesthetically pleasing poetry
in zero-shot and few-shot scenarios compared to human compositions \cite{sawicki2023bits, sawicki2023power}. The complexity and constraints in poetry generation make it a suitable scenario for multi-agent settings, given the known capacity of multi-agent systems to solve complex tasks and the potential to enhance poetry diversity by leveraging ``mixtures of multiple poets’’ \cite{yi2020mixpoet}. \newline

\noindent\textbf{Why multi-agent system and social learning?} \\
The current process of generating poetry in machine learning and NLP differs substantially from the way in which humans learn and compose poetry. Different from the paradigm of training one single poetry generation model that learns from particular datasets, human beings learn within a \textbf{social context} where they interact with others through communication, be it spoken or written, formal or non-formal \cite{jarvis2012towards}. This raises the question of whether a more human-like generation approach could improve poetry production. Multi-agent systems, with rich application\se{s} in social simulation \cite{park2023generative, chuang2023simulating}, are suitable for our goal where human behavior, such as poetry composition, can be effectively replicated through LLM-based agents that can be situated well in social networks. Furthermore, as one of the most fundamental elements for creative composition, \emph{"divergent"} thinking ability (to deviate from established norms) during the learning phase is crucial for both human and computational creativity \citet{elgammal2017can, brinkmann2023machine,wingstrom2023redefining}.
This serves as a rationale for us to adopt a social learning process that supports not only collaboration but also opposition as the main source of deviation in the process \cite{eger2016opinion, shi2019dynamics}. Moreover, unlike previous studies that focus on weak forms of opposition, such as debates and arguments, which emphasize refining differences through thought correction \cite{chan2023chateval}, our proposed social learning process aims to amplify divergence to enhance diversity. \\

{\noindent\textbf{\se{Why non-cooperative environments?}} \\
One aspect of human behavior is non-cooperative interactions in various contexts. For example, the political arena is often characterized by forms of opposition between individual parties or, in a wider context, `counter-cultures' rebelling against the establishment. Moreover, one defining property of literature / art / philosophy is to distinguish oneself from previous or contemporary `competitors'.
To name some examples, Hemingway's `iceberg' writing style differs substantially from his predecessors, \se{characterized by a more sentimental writing style} \cite{baker1972hemingway}; Impressionist artists \se{have} challenge\se{d} the standards of paintings set by the conventional art community with new contents and styles; philosopher Arthur Schopenhauer has strongly argued against the philosophy of Hegel and philosophers often form opposing groups, e.g., `Kantians', `Neoplatinists', etc \cite{janaway2002schopenhauer}. In computational social science, the terms `antagonistic', `non-cooperative' or `negative relationships' are used to describe such behavior \cite{amirkhani2022consensus}. In ML or NLP, such behavior remains rarely explored or leveraged in modeling \cite{gautier2022negotiated, Leinon-cooperative}. \newline 

\noindent\textbf{How do we utilize LLM-based agents for poetry generation?} \\ 
We build our social learning framework upon a predefined social network that governs the interactions among agents shown in Figure \ref{fig:network}. This network facilitates not only cooperative interactions (``Poets appreciate each other's work and learn from the others'') but also non-cooperative interactions (``Poets dislike each other's work and deviate from each other''). We present the learning frameworks based on two different LLMs: 1) \untrainable conversational agents (GPT-4) to answer \textit{can our framework benefit in zero-shot or few-shot settings to produce diverse poetry?} and 2) \trainable agents (GPT-2), where we investigate various training and decoding configurations such as training losses and the number of interactive agents to determine \textit{which strategy is the most effective in non-cooperative environments for poetry generation?} 
Our framework comprises three main components: (1) the social network, (2) the learning process and (3) the learning strategy. Our main contribution lies in the development of a novel learning framework that integrates existing methods to extend their scope and effectiveness.\newline

\noindent\textbf{Training-based agents in non-cooperative environments generate poetry of increasing diversity and novelty over time:} Our evaluation in Section \ref{sec:results} suggests that our framework benefits the generation process resulting in increasing diversity and novelty according to distinct and novel n-grams for \trainable agents. The generated poetry from \trainable agents also exhibits group divergence in terms of lexicons, styles and semantics per the predefined group affiliation. Analysis from Section \ref{section:analysis} also indicates that non-cooperative conditions boost diversity for \untrainable agents and the potential benefits of employing a more diverse ensemble of models with \zr{non-homogeneous} agents. But \untrainable agents in our framework do not exhibit group-based divergence of any kind. Moreover, \untrainable agents are prone to generating poetry of homogeneous styles over time.     

\section{Related work}
Our research connects to 1) \textit{the interaction of LLM-based agents} where we focus on the forms of interaction; 2) the corresponding methods to model interactions, i.e., \textit{language model ensemble and controlled text generation} (CTG) where CTG techniques are required for our use case in 3) \textit{poetry generation}. \newline

\noindent\textbf{The interaction of LLM-based agents} The form of interactions among agents can be broadly categorized as cooperative and non-cooperative. Very often, agents communicate cooperatively where the aim is to make joint decisions through collaboration such as back-and-forth communication \cite{li2024camel}, majority voting \cite{hamilton2023blind} or a combination of both \cite{zhuge2023mindstorms}. Non-cooperative interactions, though less prevalent in comparison, can enhance the quality of responses through debates or arguments among agents \cite{chan2023chateval, du2023improving}. Moreover, while most studies focus on static interaction among agents, some researchers delve into the dynamics of agents' interactions. \citet{liu2023dynamic} propose a dynamic interaction architecture where they utilize an optimization algorithm to select the best agents at inference time. Autogen \cite{wu2023autogen} also enables dynamic group chat to guide the flow of interaction among agents during ongoing conversations. Others 
focus on the output dynamics during the interaction, \se{instead}. \citet{chuang2023simulating} find that LLMs tend to align with factual information regardless of their personas and initial states which limits the simulation of
opinion dynamics using LLM-based agents. 
More recently, the dynamics of group interaction is also studied to show \se{that} incorporating chain-of-thought reasoning, detailed personas, and finetuning LLMs can enhance agents’ ability to replicate human-like group dynamics \cite{chuang2024wisdom}. Our work differs in that 1) we build a social network with group affiliations to obtain a more human-like learning process; 2) we propose a framework that involves two forms of interaction, especially with an emphasis on non-cooperative communication ; 3) we focus on the output dynamics of the generation process under finetuning for \trainable agents and consecutive prompting using detailed personas for \untrainable agents.
\newline 

\noindent\textbf{Language model ensemble and controlled text generation} 
The modeling of our framework requires 1) an ensemble of multiple language models (LMs) and 2) the generated outputs to capture general poetic styles concerning the use case of poetry generation. This involves the areas of \textit{LM ensemble} and \textit{controlled text generation (CTG).} 

\textit{LM ensemble} can be divided into 1) conversational ensemble which does not involve parameter training \cite{wang2022self}  and 2) finetuning-based ensemble i.e., neural network ensemble \cite{shazeer2016outrageously} and output ensemble \cite{dekoninck2023controlled, jiang2023llm}. Prompting-based conversational ensemble is often utilized for reasoning tasks where LLMs ensemble their own responses (i.e., self-ensemble) \cite{wang2022self, fu2022complexity}. Very recently, \citet{lu2024blending} combine multiple small conversational models in a parameter-efficient and interpretable way that outperforms ChatGPT according to their A/B test. Finetuning-based ensemble\se{s} can operate at the neural network or output level. While \textit{neural network ensemble} typically requires massive training or finetuning through extensive datasets and resources \cite{shazeer2016outrageously, jiang2024mixtral}, the mixture of smaller modules such as adapters becomes a viable solution in resource-restricted situations \cite{wang2022adamix, chronopoulou2023adaptersoup}.

\textit{CTG}, the task of generating texts subject to attributes such as emotion \cite{emosen, ruan2021emotion},  topic \cite{dathathri2019plug, wang-etal-2019-topic}, toxicity avoidance \cite{liu2021dexperts}, style \cite{belouadi-eger-2023-bygpt5, shao2021sentiment}, debiasing \cite{dinan2020queens, sheng2020towards}, etc., is also relevant to our study. CTG can operate at the training/finetuning and inference stage. Similarly to LM ensemble, finetuning with additional modules such as task-related adapters \cite{ribeiro2021structural, lin2021adapter} is also utilized to gain parameter-efficient controllability. Moreover, for CTG applications to reduce the probability of generating undesirable attributes, in addition to the standard \ce loss utilized for text generation finetuning, other loss functions are also explored. \citet{qian2022controllable} additionally include \contra loss which is crucial for the detoxification task but only partially improves the performance of the sentiment control task. \citet{zheng-etal-2023-click} also employ a \contra loss on sequence likelihood to decrease the generation probability of negative samples. In comparison, operation at the inference stage is more viable in the era of LLMs \cite{jiang2023llm, wang2023fusing, dekoninck2023controlled}. Reranking the outputs is a popular solution. For example, \citet{jiang2023llm} first rerank the complete candidate outputs from multiple LLMs and then fuse the top-K answers. Reranking the original next-token distribution during the decoding stage is also widely explored such as utilizing discriminators \cite{dathathri2019plug} or combining opinions (i.e., output logits) from (anti-)expert models \cite{liu2021dexperts, dekoninck2023controlled}. Operation during inference, especially at the decoding stage, offers strong controllability over the generated texts with less requirement on time and computational resources. However, it may cause a slight decrease in text quality \cite{dathathri2019plug}. On the other hand, operations at the training/finetuning stage preserve high-quality text with weaker controllability \cite{zhang2023survey}. 

For our use case of poetry generation, we consider both prompting- and finetuning-based ensemble methods. For the finetuning-based method, we operate jointly 1) at the training stage experimenting with standard \ce loss and (or) \contra loss to finetune with adapters for better quality texts and efficiency and 2) at the decoding stages to obtain better controllability in the non-cooperative environments. \newline

\noindent\textbf{Automatic poetry generation} 
Early attempts \se{to automatic poetry generation} mainly rely on grammatical rules \cite{oliveira2012poetryme}, statistical rules \cite{jiang2008generating, greene-etal-2010-automatic} and neural networks such RNNs \cite{zhang-lapata-2014-chinese, ghazvininejad-etal-2017-hafez, wockener-etal-2021-end}, especially RNN-based encoder-decoder architecture \cite{wang-etal-2016-chinese, lau-etal-2018-deep, Yan2016iPA}. More recent models focus on transformer-based architectures \cite{tian2021anchibert, 10.1145/3459637.3481964}. Although variants of GPT models have demonstrated outstanding performance in many NLP tasks, 
mixed opinions are observed on their ability to generate poetry. Studies such as \citet{bena2019introducing, liao2019gpt, 10.1145/3483529.3483537} finetune GPT-2 with additional components such as emotion, form and theme and result in moderate to high-quality poems according to human evaluation. \citet{kobis2021artificial} claim that \textit{zero-shot} GPT-2 can generate human-like poems where the best poem according to human selection can match human-written ones but without human preselection, machine-generated poems are easily identifiable.
\citet{wockener-etal-2021-end} point out that similar to RNN-based models, GPT-2 faces problems learning poetry-specific characteristics such as rhyme and meter. To counter such deficiencies, \citet{belouadi-eger-2023-bygpt5} propose \bygpt, an end-to-end token-free model conditioned on rhyme, meter and alliteration. The model can outperform larger models such as GPT-2, ByT5, and ChatGPT (GPT3-3.5) according to both automatic and human evaluation. They \se{also} construct a customized corpus \quatrain consisting of large-scale machine-labeled pseudo-quatrains to enlarge the finetuning dataset size. Moreover, \citet{sawicki2023power} claim that GPT-3 finetuned on 300 poems can successfully generate high-quality poems in a specific author’s style but GPT-3.5 without finetuning leads to undesirable poems written with similar styles. The findings from \citet{sawicki2023bits} also point out
that GPT-3.5 and GPT-4, without finetuning, fail to generate poetry in desired styles. Recently, interactive poetry generation has attracted the attention of researchers as it facilitates human-machine collaboration to generate poetry of more diverse styles and better quality under specific constraints \cite{zhipeng-etal-2019-jiuge, uthus-etal-2022-augmenting}. \citet{ma-etal-2023-yu} propose a post-polishing system that fine-grains GPT-2 generation based on constraints from humans. \textsc{CoPoet} from \citet{chakrabarty-etal-2022-help} finetune\se{s} pretrained T5 with <instruction, generation> pairs to enable poetry generation according to human instructions. Their study shows that finetuned T5 model is not only competitive to the larger \textsc{InstructGPT} model but also successfully collaborates with humans to produce better poems. In our study, we utilize GPT-2 as our base model as it is a good trade-off between parameter efficiency and language proficiency. Contrary to most poetry generation objectives that optimize few poetry-specific characteristics, we leverage poems with arbitrary styles where we initialize our models with randomly drawn samples from \quatrain corpus that contain pseudo poetic features. Additionally, we also explore the potential of GPT-3.5 and GPT-4 in an interactive environment enabled by a multi-agent system.  \newline

\section{Social learning framework for poetry generation}
\label{sec:sl_framework}
This section introduces our social learning framework for poetry generation. The recent development of LLMs has incentivized various attempts to simulate the social learning processes of human individuals via \llmagent \cite{li2023quantifying, chuang2023simulating, gao2023s}. 
Our framework, inspired by such processes, applies a social network-based approach to poetry generation. We investigate whether a more human-like learning process (i.e., social learning) can facilitate poetry generation. We differ from the previous studies in two aspects: 1) We base our architecture on a signed network where agents not only interact in cooperative \se{but also in} non-cooperative manners; 2) \se{w}e introduce the learning framework for \untrainable agents (GPT-3.5 and GPT-4) and \trainable agents (GPT-2). Our framework consists of three parts: \se{(1)} the social network, \se{(2)} the learning process and \se{(3)} the learning strategy. \se{We describe these below.}

\subsection{The social network}
We consider a signed social network with 
\se{$M$}
LLM-empowered agents where each link between two agents is associated with a positive or negative sign \cite{2010signed_network_media, eger2016opinion, shi2019dynamics}. We divide 
\se{the $M$}
agents into two groups as shown in Figure \ref{fig:network}. Agents within the same group are referred to as \textit{\ingroup} agents while agents from different groups are termed \textit{\outgroup} agents. We expect two types of interaction between agents based on group affiliation: 1) \ingroup agents cooperate closely with one another as `friends' (\textcolor{green}{positive sign}); 2) \outgroup agents are `foes' and they adjust their `opinions' in a non-cooperative manner (\textcolor{red}{negative sign}). We call the learning process associated with \ingroup agents \textbf{positive learning} and \outgroup associated learning is \textbf{negative learning}. Simultaneous learning 
\se{from} 
both \ingroup and \outgroup is called \textbf{joint learning} by us. 

In our application for poetry generation, the agents are pretrained LLMs. 
We view the LLMs as different poets belonging to two groups. The \ingroup poets appreciate each other’s work and aim to learn from their styles. Conversely, the \outgroup poets dislike each other’s work and aim to differentiate their works.

\subsection{The learning process}\label{subsec:learning}
\begin{table}[!htp]\centering
\small
\begin{tabular}{m{1.2cm}m{9cm}}\toprule
Variable &Definition \\\midrule
$a_1, a_2,\ldots$ &agents belong to group A \\
$b_1, b_2,\ldots$ &agents belong to group B \\
M & the total number of agents \\
$\aij$ &the target agent $\aij \in \{a_1, b_1, a_2, b_2, \ldots\}$ where $i \in \{1, 2, \ldots, M\}$ \\
$\aij, \aijin, \aijout$ &the agent tuple: (the target agent, the \ingroup agents of the target agent, the \outgroup agents). E.g., ($a_1, [a_2, a_3], [b_1, b_2]$) \\
$\pa, \pap, \pan$& the conditional probability distribution for the next token of agent $\aij$, agents $*^+ \in \aijin$ and agents $*^- \in \aijout$  \\
N &the total number of generated poems per iteration per agent \\
T &the total number of iterations \\
t &the iteration number $t \in \{1,2, \ldots, T\}$ \\
$o^{\aij}$ &a generated poem by agent $\aij$ \\
$O_{t}^{\aij}$ & the set of poems generated by agent $\aij$ at iteration t \\
$S_t$ &the set of all generated poems at iteration $t$ \\
$\fgenerate$ &a generation function of agent tuple $(\aij, \aijout, \aijin)$ \\
$\fupdate$ &an updating function based on the latest generated outputs $S_t$ and $S_{t-1}$ \\
$t_g$ & the generation time at the decoding stage\\
$x_{t_g}$ & the token at generation time $t_g$  \\
$\vect{x_{t_g}}$ & the input sequence at generation time $t_g$ \\
\#$\mathcal{A}$ & the number of interactive agents at the decoding stage \\
\bottomrule
\end{tabular}
\caption{Notations}\label{tab:not}
\end{table}
\begin{algorithm}
\caption{Social learning process for poetry generation}
\label{algo:learningprocess}
\algrule
\For{$t \gets 1$ \KwTo $T$}{
    Initialize an empty set $S_t$ to store the generated poems at iteration $t$\;
    \ForEach{agent $\aij$ \tcp*[f]{$i \in \{1, 2, \ldots, M\}$}}{
        $O_{t}^{\aij} \gets \fgenerate(\aij, \aijout, \aijin)$ to generate $N$ poems\;
        Add $O_{t}^{\aij}$ to $S_t$\; 
    }
    \ForEach{agent $\aij$ \tcp*[f]{$i \in \{1, 2, \ldots, M\}$}}{
        \eIf{$t > 1$}{
            $\aij  \gets \fupdate(S_t, S_{t-1})$\;
        }{
            $\aij  \gets \fupdate(S_t)$\;
        }
    }
}
\algrule
\end{algorithm}
Now, we describe the learning process among agents based on the social network shown in Figure \ref{fig:network} and Algorithm \ref{algo:learningprocess}. We summarize all the notations mentioned below in Table \ref{tab:not}. The learning process represents the high-level communication procedure among agents. We begin with pretrained LLM-based agents $a_1, a_2,\ldots$ belonging to group $A$ and $b_1, b_2, \ldots$ belonging to group $B$ (see Section \ref{sec:setup} for details of agent initialization). We further divide the learning process into two phases: the \pup phase and the \pgen phase, defining the 
learning strategy $(\fupdate, \fgenerate)$. The two functions jointly organize the positive, negative and joint learning process. $\fgenerate$ is a generation function that outputs poems $O$. $\fupdate$ is a learning function that equips agents with the latest knowledge based on the generated poems. We denote
\se{an} agent as $\aij$ where $i \in \{1,2, \ldots, M\}$. 
The \outgroup agents for $\aij$ are referred to as $\aijout$ and the \ingroup agents as $\aijin$. The generation function is thus $\fgenerate(\aij,\aijout, \aijin)$.

At each iteration $t$, the learning process starts with the \pgen phase. Each agent $\aij$ generates a set of $N$ poems through function $\fgenerate(\aij,\aijout, \aijin)$. We iterate over all agents and collect a set of poems $S_t$ which consists of all generated poems $O_t$ from the current iteration $t$. Then, we let agent $\aij$ update their knowledge and learn cooperatively, non-cooperatively or jointly from each other based on poems from the current iteration $S_t$ and the previous iteration $S_{t-1}$.\footnote{We utilize the generation from the current and previous iteration to expand the dataset size of the finetuning step where the agent can update to the latest knowledge and prevent potential catastrophic forgetting \cite{biesialska2020continual}.} We denote the update function as $\fupdate(S_t, S_{t-1})$. We iterate over all agents $\aij$ for $i \in \{1,2, \ldots, M\}$ until 
\se{all agents have been updated.} 
We conduct the learning process $T$ times. 

\subsection{The learning strategy}\label{subsec:strategy}
As mentioned in Section \ref{subsec:learning}, the learning process involves positive, negative and joint learning which operates at the \pgen and the \pup phase. We now discuss the detailed learning strategy for \trainable agents and \untrainable agents. For \trainable agents, the learning strategy contains finetuning strategies for the \pup phase \& decoding strategies for the \pgen phase. For \untrainable agents, both phases are conducted via prompting. Table \ref{tab:learning_strategy} summarizes the learning strategies for both types of agents. 
\begin{table*}[!htp]\centering
\small
\begin{tabular}{lllll}
\toprule
Type of agents & Strategy & Positive learning & Negative learning & Joint learning  \\
 & & ($\aij, \aijin$) & ($\aij, \aijout$) & ($\aij, \aijin, \aijout$) \\
\midrule
\multirow{3}{*}{\trainable} & decoding & - & $\pa$, $\pan$ & $\pa, \pap, \pan$ \\
&\multirow{2}{*}{finetuning} & \multirow{2}{*}{\lce} & \multirow{2}{*}{-}
& 1) \lcl \\
& &&& 2) conditioned \lce \\
\midrule
\untrainable &prompting &\multicolumn{2}{c}{\pchain} & \pjoint \\
\bottomrule
\end{tabular}
\caption{Learning strategies for \trainable agents and \untrainable agents. \lce and \lcl represent \ce loss and \contra loss.  $\aij, \aijin, \aijout$ denotes the target agent, the \ingroup agent and the \outgroup agent of the target agent $\aij$. $\pa, \pap,  \pan$ are the conditional probability distribution for the next token of agent $\aij$, agents $*^+ \in \aijin$ and agents $*^- \in \aijout$ defined in Equation (\ref{equ:decode}). }\label{tab:learning_strategy}
\end{table*} 
\subsubsection{Training-based agents}
We first introduce the decoding strategy for the \pgen phase where we utilize reranking techniques. We then detail the finetuning strategy for the \pup phase where we employ the \contra loss and the standard \ce loss. \newline

\noindent\textbf{Decoding strategy at the \pgen phase}

We adopt the \dex framework, where models learn through a comparison and contrast mechanism \cite{liu2021dexperts}. By reranking the probability distribution of the next token, the target agent $\aij$ generates the next token by jointly considering the probability distribution of the target agent $\aij$ itself, its \ingroup agents $\aijin$ and \outgroup agents $\aijout$. At the decoding stage, the number of interactive agents involved can vary depending on the specific subset chosen from the sets $\aijin$ and $\aijout$, denoted as $\aijin_{\#}$ and $\aijout_{\#}$. The number of interactive agents is thus \#$\mathcal{A}$.\footnote{\#$\mathcal{A}= \rvert \aijin_{\#} \rvert + \rvert \aijout_{\#} \rvert$} This flexibility allows the system to dynamically adjust the number of interactive agents participating in the decoding process, offering adaptability based on the requirements or constraints of the task. The detailed formulation of the decoding strategy is shown below.  

Given the input sequence at generation time $t_g$ ($g$ indicates the \textit{generation} stage), 
denoted as $\vect{x_{<t_g}}$, we predict the next token $x_{t_g}$ generated by the target agent $\aij$ through combining the outputs from $\aijin_{\#}$ and $\aijout_{\#}$. We first obtain the conditional logit scores of all models denoted by $l_{\aij}(x_{t_g}|\vect{x_{<t_g}}), l_{*^+}(x_{t_g}|\vect{x_{<t_g}}), l_{*^-}(x_{t_g}|\vect{x_{<t_g}})$, where $*^+ \in \aijin_{\#}$ is an agent belonging to the interactive \ingroup agents $\aijin_{\#}$ and $*^- \in \aijout_{\#}$; $l_{*}(x_{t_g}|\vect{x_{<t_g}}) \in \mathbb{R}^{|\mathcal{V}|}$ and $\mathcal{V}$ is the vocabulary. The probability distribution of the next token over the vocabulary $\mathcal{V}$ is $P_{*}({x_{t_g}|\vect{x_{<t_g}}}) = \text{softmax}[l_{*}(x_{t_g}|\vect{x_{<t_g}})]$. The probability distribution of the next token $\hat{P}_{\aij}({x_{t_g}|\vect{x_{<t_g}}})$ is thus given by 
\begin{equation}
\label{equ:decode}
\begin{split}
    \hat{P}_{\aij}({x_{t_g}|\vect{x_{<t_g}}}) = & \text{softmax}\{l_{\aij}(x_{t_g}|\vect{x_{<t_g}}) + \\
    & \alpha [\frac{\sum_{\aijin_{\#}}{l_{*^+}(x_{t_g}|\vect{x_{<t_g}})}}{\rvert \aijin_{\#} \rvert}-\frac{\sum_{\aijout_{\#}}{l_{*^-}(x_{t_g}|\vect{x_{<t_g}})}}{\rvert \aijout_{\#} \rvert}]\}
\end{split}
\end{equation} 
The next token $x_{t_g}$ is assigned with high probability if the probability is high under both $\pa$ and $\pap$ and low under $\pan$. Moreover, if we replace $\pap$ with $\pa$, the process considers \outgroup agents $\aijout$ only which solely models negative learning. This decoding strategy can model both negative learning ($\pa, \pan$) and joint learning ($\pa,\pan, \pap$) as shown in Table \ref{tab:learning_strategy}. 
\newline

\noindent\textbf{Finetuning strategy at the \pup phase}
We discuss the finetuning strategies based on the learning relationships, i.e., positive and joint learning as shown in Table \ref{tab:learning_strategy}. 
\begin{itemize}
    \item \textit{Positive learning}: we utilize the conventional finetuning method with \ce loss (\lce) to finetune agent $\aij$ with poems $o \in O_{\aij} \cup O_{\aijin}$ (poems generated from $\aij$ and $\aijin$). The loss function for the $j^{th}$ poem $o_j$ in a mini-batch is thus 
    \begin{equation}
\mathcal{L}_{\text{CE}}(\aij, o_j) = - \sum_{t_g=1}^{\mathcal{T}} \log(P_{\aij} (x_{t_g} | \vect{x_{<t_g}}))
    \end{equation} where $\mathcal{T}$ denotes the number of tokens for the poem $o_j$. $P_{\aij} (x_{t_g} | \vect{x_{<t_g}})$ is the conditional distribution of the token at time $t_g$ for poem $o_j$ given the previous sequence $\vect{x_{<t_g}}$.  
    \item \textit{Joint learning with \contra loss}: Our social network design well suits the setting for \contra learning where poems from \ingroup agents are positive samples and poems from \outgroup agents are negative samples. We adopt \contra learning to pull closer the semantic representation of \ingroup samples and push apart that of \outgroup samples. We implement the \contra loss \textsc{SimCSE} proposed by \citet{gao2021simcse}. For a mini-batch containing samples from $\aij, \aijin, \aijout$, let $(o_j^{\aij}, o_j^{\aijin}, o_j^{\aijout})$ denote the $j^{th}$ paired triple and $(\vect{h_j}, \vect{h_j}^{+}, \vect{h_j}^{-})$ be its representation. The \contra loss is thus 
    \begin{equation}
       \mathcal{L}_{\text{CL}} (\aij, (\vect{h_j}, \vect{h_j}^{+}, \vect{h_j}^{-}) ) =  - \log \frac{e^{\text{sim}(h_j, h_j^+)/\tau}}
{\sum_{k=1}^{Q} \left( e^{\text{sim}(h_j, h_k^+)/\tau} + e^{\text{sim}(h_j, h_k^-)/\tau} \right)}
    \end{equation} where $Q$ is the size of mini-batch, $\tau$ is the temperature and $sim(h_1, h_2)$ is the cosine similarity $\frac{h_1^\mathsf{T}h_2}{\| h_1 \| \cdot \| h_2 \|}$. Here\se{,} we experiment with \contra loss \se{(a)} alone and \se{(b)} jointly with \lce for \ingroup poems.
    \item \textit{Joint learning with conditioned \ce loss}: We utilize the style constraints \cite{belouadi-eger-2023-bygpt5} using conditions <positive> and <negative> for poems generated by \ingroup agents and \outgroup agents. We then finetune the agent $\aij$ with \ce loss.   
\end{itemize}
\subsubsection{Prompting-based agents} 
The learning strategy for \untrainable agents relies on prompting. Our prompts are constructed with three modules: 1) a profile module, which defines the role of $\aij$; 2) a memory module, which stores the generated poems; and 3) an action module, which completes the generation task. Table \ref{apdx:prompt} in the appendix contains the prompts for both prompting strategies. For \untrainable agents, the \pup and \pgen phases do not function in isolation. $\fupdate$ updates the profile module during prompting based on the generated poems from previous iterations. Similar to the design of \trainable agents, we can update the knowledge of agents based on different learning relationships: 
\begin{itemize}
    \item The \pchain strategy for isolated positive \& negative learning: For \pchain, we update the knowledge of $\aij$ based on its relationships with other agents in a separate manner. At iteration $t$, we first update the profile of $\aij$ with poems generated from $\aijin$ at time $t-1$. The process is denoted as $\fupdate(\aij, \aijin)$. $\aij$ thus generates a poem $o^{\aij}$ based on the positive learning results (example prompt: ``Please read the poems from your friends carefully and compose similarly to your friend.''). We then update the profile of $\aij$ with the poem $o^{\aij}$ and a poem $o^{\aijout}$ sampled from the previous iteration $t-1$. This process is denoted as  $\fupdate(\aij, \aijout)$. $\aij$ thus recomposes the poem $o^{\aij}$ based on the negative learning results (example prompt: ``Please rewrite your poem to compose dissimilarly to your foe.''). 
    \item Joint prompting for joint learning: Joint prompting updates the profile with poems generated by $\aijin$ and $\aijout$ at the same time, denoted as $\fupdate(\aij, \aijout, \aijin)$. 
\end{itemize}

\section{Experiments}
\label{sec:setup}
\subsection{Agent initialization}
\begin{table}[h]\centering
\small
\begin{tabular}{m{6cm}m{1.2cm}m{1.2cm}m{1.2cm}}
\toprule
Instance & Rhyme & Meter & alliteration\\
\midrule
Who hath such beauty seen \newline
In one that changeth so? \newline
Or where one's love so constant been, \newline 
Who ever saw such woe? & ABAB & iambus	& 0.11 \\
\arrayrulecolor{black! 10}\midrule
Would rather seek occasion to discover \newline
How little pitiful and how much unkind, \newline
They other not so worthy beauties find. \newline
O, I not so! but seek with humble prayer  & ABBC & iambus & 0.05 \\
\arrayrulecolor{black! 10}\midrule
Of pearl and gold, to bind her hands; \newline
Tell her, if she struggle still, \newline
I have myrtle rods at will, \newline
For to tame, though not to kill. & ABBB & iambus & 0.10 \\
\arrayrulecolor{black! 100}\bottomrule \\
\end{tabular}
\caption{Instances from \quatrain corpus.}
\label{tab:quatrain}
\end{table}
\noindent\textbf{Initialization for \trainable agents}
We choose $M=4$ for our initial experiments. We first pretrain GPT-2 (medium) with \se{a subset of the} random \quatrain corpus of size 123K (nearly 1/6 of \quatrain corpus). \quatrain consists of machine-labeled pseudo-quatrains that are consecutive sequences with four lines extracted from real human poems as shown in Table \ref{tab:quatrain}. The poems follow poetic characteristics, i.e., rhyme, meter and alliteration. We pre-select the \quatrain instances excluding those that are semantically similar ($>0.7$) with each other based on pairwise cosine similarity of sentence embedding\se{s} calculated using \sbert \cite{reimers-gurevych-2019-sentence}. We first pretrain GPT-2 further with 720 training steps using the pre-selected \quatrain dataset.
More details of the pretraining and the loss curve are shown in Section \ref{apdx:para}. We then finetune the pretrained model with four randomly selected \se{subsets} of \se{size} 7.5k \se{from our 123K subcorpus} to obtain different initializations of four agents. The initialization step prepares models with a preliminary understanding of poetic structures and characteristics essential for subsequent learning phases.  \newline

\noindent\textbf{Initialization for \untrainable agents}
For \untrainable agents, we randomly sample \quatrain instances and initialize the agent with \pchain and \pjoint under the predefined profile shown in Table \ref{tab:pchain} and Table \ref{tab:pjoint}. 

\subsection{Experimental setup}
\begin{table} 
\centering
\small
\begin{tabular}{m{1cm}m{0.7cm}m{1.8cm}m{0.2cm}m{0.2cm}m{1.2cm}m{8cm} }
\toprule
\multirow{2}{*}{RQ1} &\multicolumn{2}{c}{Para\_decoding} &\multicolumn{3}{c}{Para\_finetuning} &\multirow{2}{*}{Description} \\
\cmidrule(r){2-3}\cmidrule(l){4-6}
& \#$\mathcal{A}$ & $\alpha$ & \lce & \lcl & \textit{conditioned} & \\
\midrule
\#$\mathcal{A}$ during decoding &\{2,3,4\} & 2 &X  & & &The number of agents involved during decoding is varied. \newline 1) \#$\mathcal{A}$ = 2: negative decoding + positive finetuning \newline 2) \#$\mathcal{A}$ > 2: joint decoding + positive finetuning  \\
\arrayrulecolor{black! 10}\midrule
$\alpha$ &2 & \{0, 1, 1.5, 2, 2.5\} &X  & & & The scaling parameter $\alpha$ during decoding is varied. \newline 
$\alpha > 0$: negative decoding + positive finetuning \newline
$\alpha = 0$: positive finetuning only, `echo chamber' \\
\arrayrulecolor{black! 10}\midrule 
\multirow{5}{*}{\parbox{1cm}{finetuning strategy}} &\multirow{5}{*}{2} &\multirow{5}{*}{2} &X & & &\multirow{5}{*}{\parbox{9cm}{Different training loss applied for $\alpha$ = 2 (with negative decoding). \newline 1) \lce alone: positive training with negative decoding \newline 2) \lcl or \textit{conditioned}: joint training with negative decoding}} \\
& & &X  & & & \\
& & &  &X & & \\
& & &X  &X & & \\
& & &X  & &X & \\
\arrayrulecolor{black! 100}\bottomrule \\
\end{tabular}
\caption{Experimental setup for \trainable agents. Para\_decoding and Para\_finetuning represent parameters during the decoding and finetuning stage. \#$\mathcal{A}$ is the number of agents. $\alpha$ is the scaling parameter in Equation (\ref{equ:decode}).}
\label{tab:setup}
\end{table}
For \trainable agents, we design experiments to explore how the parameters from the finetuning stage (i.e., loss functions) and the decoding stage (number of agents and the scaling parameter) affect the dynamics of generation. We summarize the detailed setup in Table \ref{tab:setup}. 

For \untrainable agents, we design experiments with different prompting strategies, i.e., \pchain and \pjoint using both GPT-3.5 (gpt-3.5-turbo) and GPT-4 (gpt-4-turbo).

\subsection{Evaluation}\label{sec:eval}
We first conduct automatic evaluation where we study the generation dynamics of our framework from lexical perspectives. We study lexical novelty and diversity, as novelty and diversity are crucial indicators for creative tasks such as poetry generation. We then study the dynamics from semantic (semantic similarity) perspectives. Lastly, we evaluate the poems qualitatively where we directly compare the generated poetry. \newline

\noindent\textbf{Metric for lexical diversity and novelty.} We measure the lexical diversity using the percentage of distinct uni-grams (\textit{distinct-1}) and bi-grams (\textit{distinct-2}) following the definition by \citet{su2022a, tevet-berant-2021-evaluating}. The formulation is given as: 
    $\frac{\text{unique n-grams}(O)}{\text{total n-grams}(O)}$,
where $O$ is the set of generated poems to be evaluated. We adopt the measure of novelty (\textit{novelty-1} and \textit{novelty-2}) by \citet{mccoy-etal-2023-much, composelikehuman} where we calculate the number of new uni-/bi-grams that do not stem from the pretraining set and rescale them with the total number of generated tokens. Thus, novelty reflects the lexical difference between the generated poems and the training set, while diversity indicates the token variety among the generated poems.  \newline

\noindent\textbf{Metric for group-based semantic similarity.}
Following our group affiliation, we examine the group dynamics of the agents by their semantics. For any pair of poems sampled from the same iteration $t$, we calculate the \textit{semantic similarity} of the paired instances by computing the cosine similarity of the embeddings retrieved from \sbert \cite{reimers-gurevych-2019-sentence}. We then aggregate the similarity scores per iteration by their group affiliation (i.e., \ingroup and \outgroup) defined in Figure \ref{fig:network}. \newline
\section{Experiment results}\label{sec:results}

\subsection{Automatic Evaluation: the generation dynamics of agents}
We generate 400 poems using the same set of decoding parameters for \trainable agents or the same prompt templates for \untrainable agents from each agent for every iteration.\footnote{See Section \ref{apdx:para} in the appendix for more details on parameter setting.} In total, we obtain more than 96k generated poems. We report the lexical diversity (\textit{distinct-1} and \textit{distinct-2}),  novelty (\textit{novelty-1} and \textit{novelty-2}), and group-based semantic similarity defined in Section \ref{sec:eval}.

Our main findings are: 1) according to lexical level comparison (i.e., distinct and novel uni-/bi-grams), our framework benefits \trainable agents resulting in increasing diversity and a higher level of novelty; 2) according to pairwise semantic similarity averaged per group affiliation, we observe group divergence in semantics for \trainable agents, especially \outgroup divergence due to operation at the decoding stage; 3) \untrainable agents generate poems of more diverse lexicons at $t=1$ but they tend to output poems of homogenous styles over time.\footnote{Due to resource limit, we do not conduct multiple runs for all experiments. We study the stability of our statistics in Section \ref{sec:stability}. This section is based on one single run of experiment\se{s}.}   

\subsubsection{Increasing diversity and novelty according to distinct and novel n-grams for \trainable agents} We compute \textit{distinct-1/2} and \textit{novelty-1/2} with all generated poems and average over all agents for every iteration $t$. The results are shown in Table \ref{tab:diveristy} and Figure \ref{fig:diversity}. \newline

\noindent\textbf{RQ1: how do different learning strategies affect the diversity and novelty of \trainable agents?}
\begin{figure}
    \begin{subfigure}{0.32\textwidth}
    \centering
    \includegraphics[width=1\textwidth,height=4.5cm]{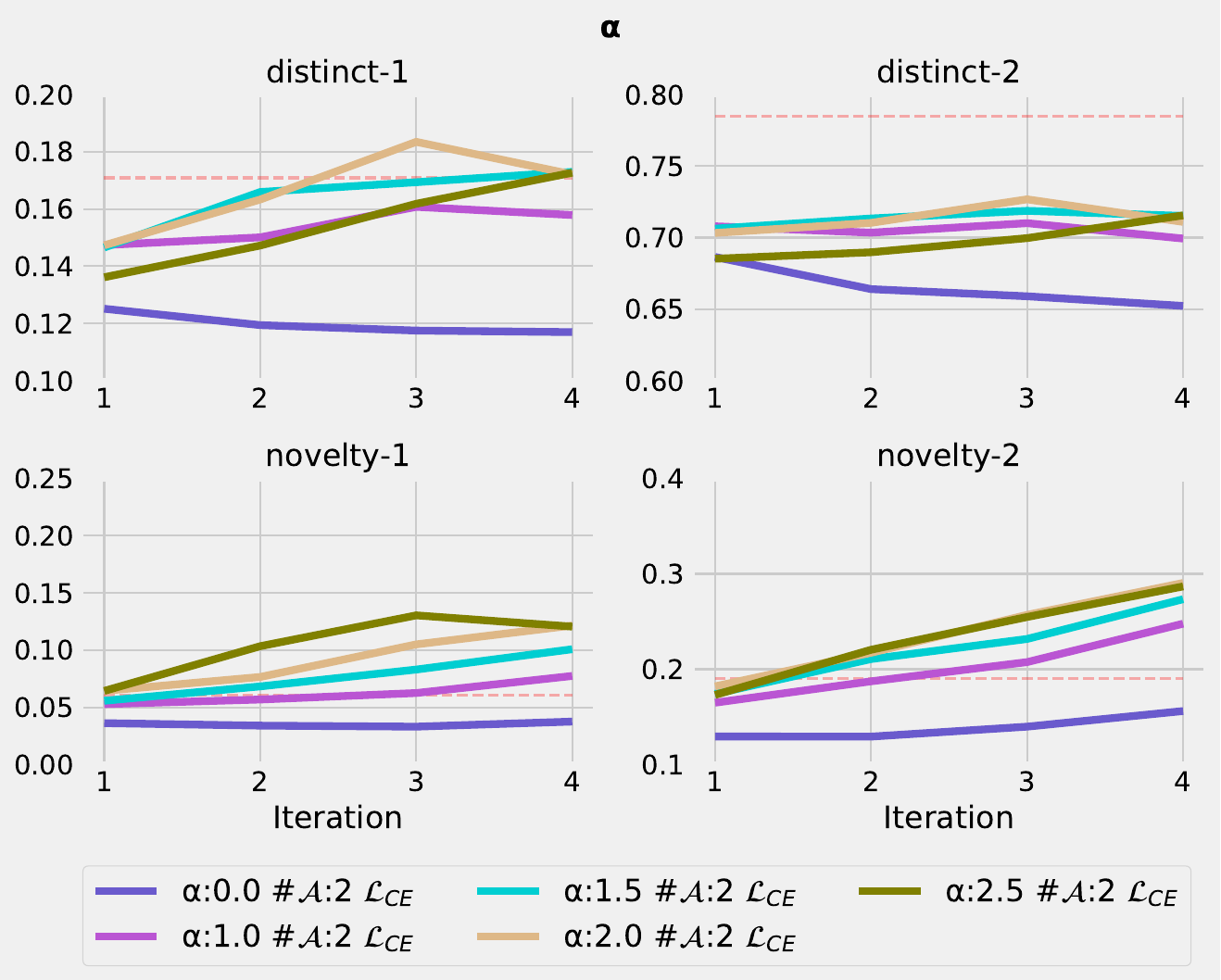}
    \caption{}\label{fig:diveristy_alpha}
\end{subfigure}
\hfill
\begin{subfigure}{0.32\textwidth}
    \centering
    \includegraphics[width=1\textwidth, height=4.5cm, ]{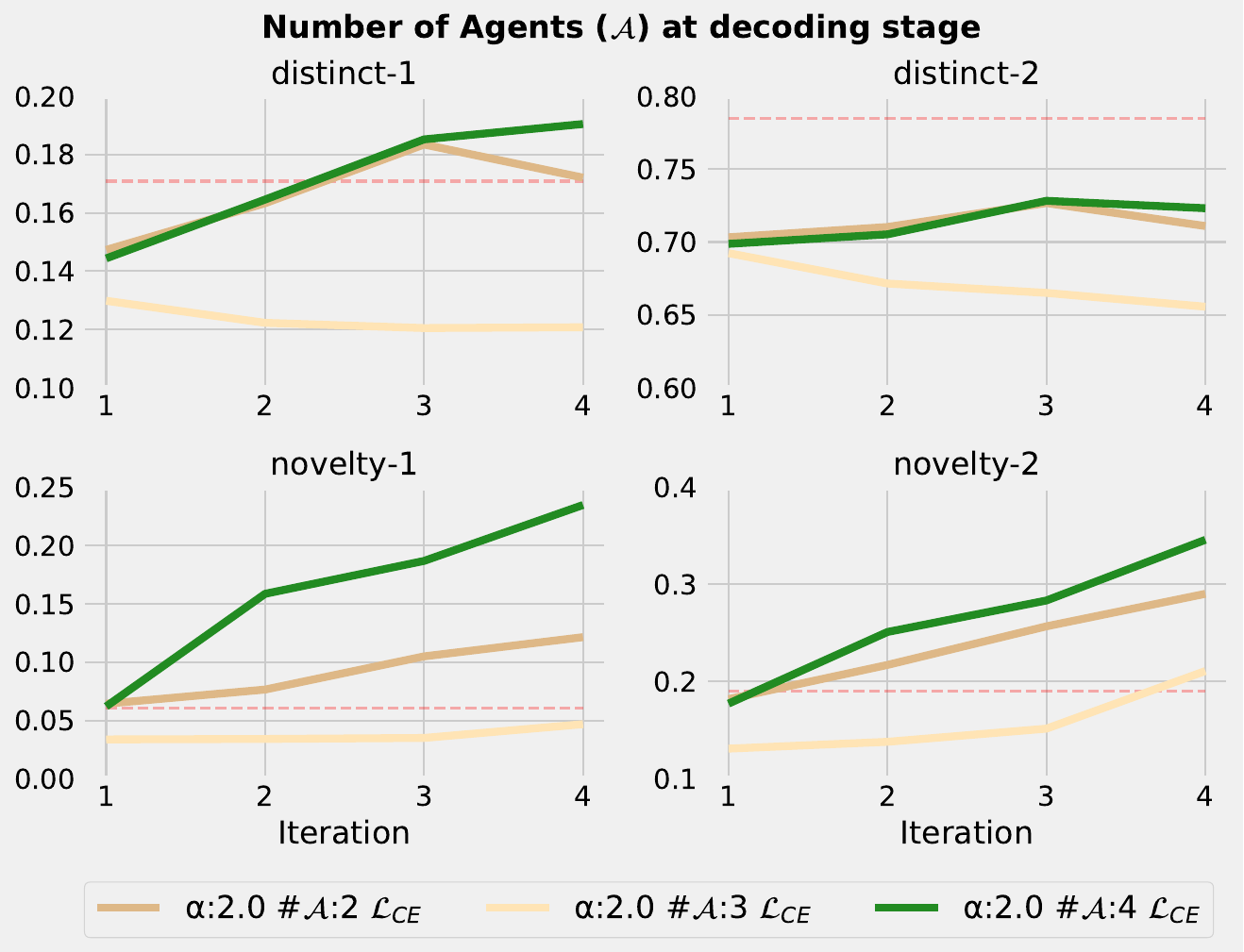}
    \caption{}\label{fig:diveristy_nagent}
\end{subfigure}
\hfill
    \begin{subfigure}{0.32\textwidth}
    \centering
    \includegraphics[width=1\textwidth, height=4.5cm, ]{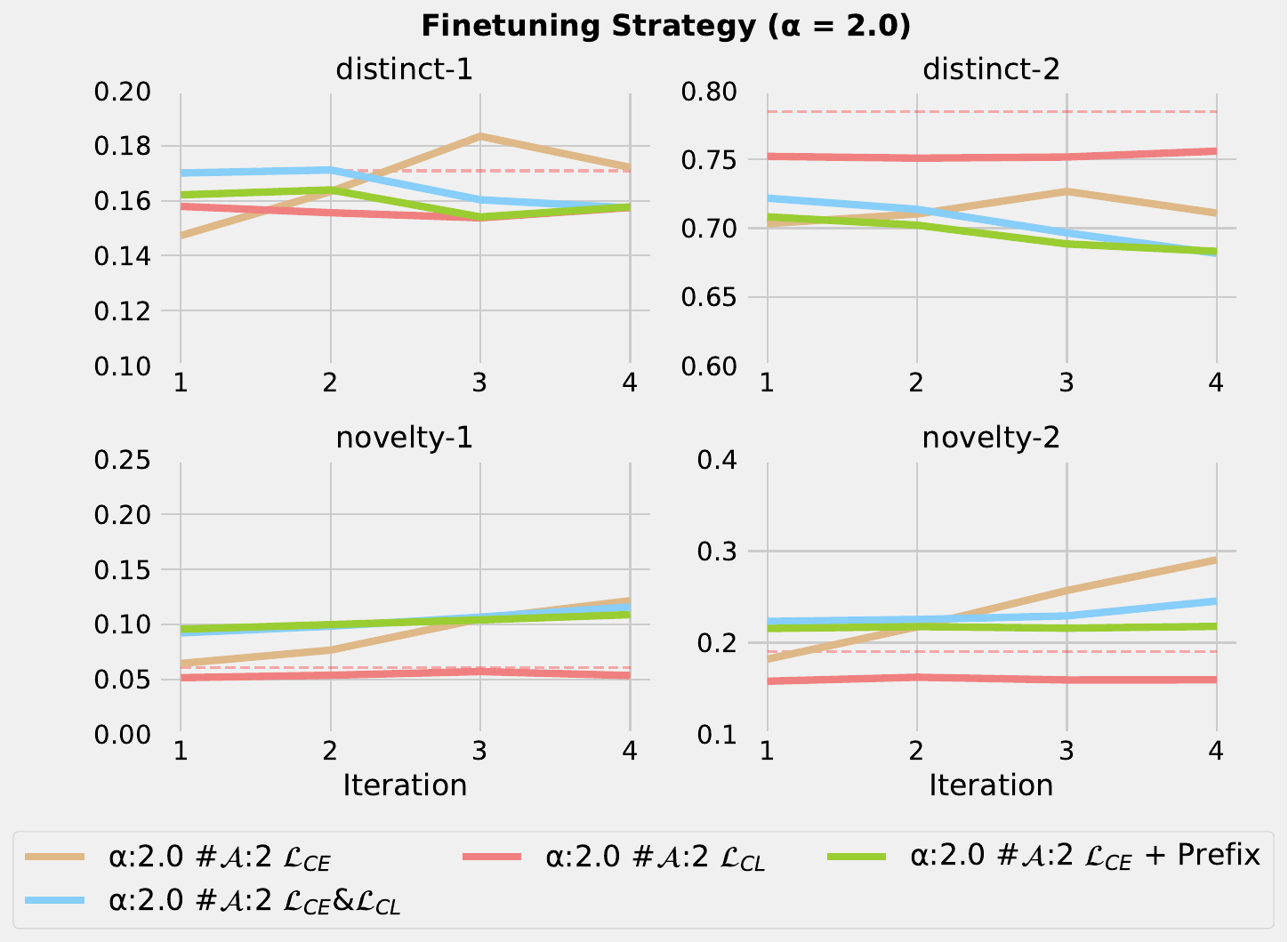}
    \caption{}\label{fig:diveristy_loss}
\end{subfigure}
\caption{Dynamics of agent diversity and novelty over varying training parameters. The degree of diversity is measured by the percentage of distinct uni-grams (\textit{distinct-1}) and bi-grams (\textit{distinct-2}) in the generated poems. The degree of novelty is measured by the number of novel uni-grams (\textit{novelty-1}) and bi-grams (\textit{novelty-2}) in the generated poems compared to that in training data scaled by the total number of generated tokens. 
(a) The effect of scaling parameter $\alpha$ in Equation (\ref{equ:decode}). (b) The effect of the number of interactive agents \#$\mathcal{A}$ during the decoding stage. (c) The effect of finetuning strategies: \lce and \lcl indicate \ce loss and \contra loss. Prefix refers to the conditioned finetuning. The horizontal red dashed line indicates the state of initial agents at iteration 0.}\label{fig:diversity}
\end{figure}
\begin{table}
\centering
\small
\begin{tabular}{lllllll}
\toprule
$\alpha$ & \#$\mathcal{A}$ & $\mathcal{L}$ & distinct-1 & distinct-2 & novelty-1 & novelty-2 \\
\midrule
\multicolumn{7}{c}{\textit{varying $\alpha$}} \\
0 &2 &\lce &0.120 &0.665 &0.035 &0.139 \\
1 &2 &\lce &0.154 &0.705 &0.062 &0.202 \\
1.5 &2 &\lce &0.164 &\textbf{0.713} &0.077 &0.222 \\
2 &2 &\lce &\textbf{0.167} &\textbf{0.713} &0.092 &\textbf{0.237} \\
2.5 &2 &\lce &0.154 &0.698 &\textbf{0.105} &0.234 \\
\multicolumn{7}{c}{\textit{varying \#$\mathcal{A}$}} \\
2 &2 &\lce &0.167 &0.713 &0.092 &0.237 \\
2 &3 &\lce &0.123 &0.671 &0.037 &0.158 \\
2 &4 &\lce &\textbf{0.171 }&\textbf{0.714} &\textbf{0.161} &\textbf{0.264} \\
\multicolumn{7}{c}{\textit{varying training loss}} \\
2 &2 &\lce &\textbf{0.167} &0.713 &0.092 &\textbf{0.237} \\
2 &2 &\lce + \lcl &0.165 &0.703 &\textbf{0.103} &0.231 \\
2 &2 &\lcl &0.156 &\textbf{0.753 }&0.054 &0.160 \\
2 &2 &\lce + prefix &0.159 &0.696 &0.102 &0.217 \\
\arrayrulecolor{black! 10}\midrule
\multicolumn{3}{c}{Initialization} & 0.171 & 0.785 & 0.061 & 0.190 \\
\arrayrulecolor{black}\bottomrule \\
\end{tabular}
\caption{Diversity and novelty results in aggregative mean for \trainable agents. \textit{Distinct-1} and \textit{distinct-2} are the percentage of distinct uni-/bi-grams. \textit{Novelty-1} and \textit{novelty-2} reflect the number of new uni-/bi-grams that do not appear in the training set rescaled by the total number of generated tokens. The highest value in each experimental setting is highlighted in \textbf{bold}. $\alpha$ represents the decoding scaling parameter; \#$\mathcal{A}$ is the number of interactive agents at decoding stage; $\mathcal{L}$ represents the loss function during finetuning. \textit{Initialization} indicates the states of initial agents at iteration 0.}\label{tab:diveristy}
\end{table}

Figure \ref{fig:diversity} shows the dynamics of agent diversity and novelty under varying training parameters and Table \ref{tab:diveristy} shows the results of different experimental setups for diversity and novelty averaged over all iterations. We observe that:
\begin{itemize}
    \item the effect of negative decoding strategy with the scaling parameter $\alpha$.
    \begin{itemize}
        \item \textbf{\textit{Diversity}} Negative decoding combined with positive finetuning ($\alpha$ >0, \lce) strategy leads to increasing diversity over time though the level of diversity is below the initial state at $t=0$. Aggregatively, results from Table \ref{tab:diveristy} (\textit{varying $\alpha$}) suggest that compared to the case without negative decoding (i.e., $\alpha=0$), negative decoding strategy under varying $\alpha$ ranging from 1 to 2.5 yields 3.4 to 4.7 percentage points (pp) increase in \textit{distinct-1} and 3.3 to 4.8 pp increase in diversity measured by \textit{distinct-2}. Dynamically, the results from Figure \ref{fig:diveristy_alpha} suggest that the lexical diversity of generated poems with negative decoding depicts an increasing trend from $t=1$ to $t=4$ for all $\alpha>0$ measured by both \textit{distinct-1} (with a maximum increase of 3.7 pp) and \textit{distinct-2} (with a maximum increase of 3.0 pp) while for $\alpha=0$ (i.e., without negative decoding), both diversity measures decrease slightly. Worth noting is that both \textit{distinct-1} and \textit{distinct-2} are below the diversity level measured at $t=0$ (shown as the red dashed line in Figure \ref{fig:diveristy_alpha} and the last row in Table \ref{tab:diveristy}), especially \textit{distinct-2}.   
        \item \textbf{\textit{Novelty}} Negative decoding combined with positive finetuning ($\alpha$ >0, \lce) strategy boosts novelty over time resulting in more novel generation compared to the initial state at $t=0$. The last two columns in Table \ref{tab:diveristy} show that the negative decoding strategy, i.e., $\alpha>0$, can boost novelty in the aggregative mean by a maximum of 7.0 pp in \textit{novelty-1} and a 9.8 pp increase in \textit{novelty-2} compared to the case without negative decoding, i.e., $\alpha=0$. Dynamically, results from Figure \ref{fig:diveristy_alpha} suggest a sharper increase over iterations for all $\alpha>0$ measured by both \textit{novelty-1} (with a maximum increase of 5.6 pp) and \textit{novelty-2} (with a maximum increase of 11.3 pp) compared to the results for $\alpha=0$.
    \end{itemize}
    \item the effect of the number of agents (\#$\mathcal{A}$) involved at the decoding stage 
    \begin{itemize}
    \item \textbf{\textit{Diversity}} As shown in Table \ref{tab:diveristy}, $\#\mathcal{A}=4$ yields the highest diversity level according to \textit{distinct-1} and \textit{distinct-2} with $\#\mathcal{A}=2$ achieving similar performance. Dynamically, diversity increases over iteration for paired agents ($\#\mathcal{A}=2$ or $4$) at the decoding stage. However, for $\#\mathcal{A}=3$, we observe a decreasing trend in diversity with much lower level of diversity compared to the case for $\#\mathcal{A}=2$ or $4$.  
    \item \textbf{\textit{Novelty}} We observe a greater gain for novelty at $\#\mathcal{A}=4$. This is evident: 1) Table \ref{tab:diveristy} shows 6.9 pp increase in aggregative mean for $\#\mathcal{A}=4$ compared to $\#\mathcal{A}=2$; 2) Figure \ref{fig:diveristy_nagent} indicates a shaper increasing trend at $\#\mathcal{A}=4$ especially for \textit{novelty-1}. Both \textit{novelty-1} and \textit{novelty-2} are above the initial state at iteration 0 which suggests a boost in novelty over all time. However, we observe less novelty for $\#\mathcal{A}=3$, which is similar to the case for diversity. 
     \end{itemize}
    \item the effect of finetuning strategy. 
    The decoding parameters \#$\mathcal{A}$ and $\alpha$ are fixed and we experiment with varying finetuning losses. As suggested by Figure \ref{fig:diveristy_loss}, the most effective finetuning strategy according to the dynamics of diversity and novelty is \lce (i.e., positive finetuning using the \ce loss) which presents an observable upward trend. Finetuning using \lcl (i.e., joint finetuning using \contra loss) yields slightly better diversity according to \textit{distinct-2}. We also observe minor improvement in novelty for strategy \lcl $+$ \lce (i.e., joint finetuning using both losses) in aggregative mean shown in Table \ref{tab:diveristy}. However, dynamically we do not spot any increase over time for both cases. Conditioned finetuning (i.e., Prefix) also fails to bring improvements. 
\end{itemize}
To sum up, our framework can lead to increasing diversity and a higher level of novelty: 1) negative decoding combined with positive finetuning ($\alpha$ >0, \lce) is the most effective combination of decoding and finetuning strategies; 2) the increase in an even number of agents can improve the results, especially for novelty; 3) in our experiment, positive finetuning (i.e., finetuning using \ce loss alone) is more effective overall both in aggregative mean and dynamically compared to other finetuning strategies.  \newline

\noindent\textbf{RQ2: how do different prompting strategies affect the diversity of \untrainable agents?}
\begin{figure}[!htb]
    \begin{minipage}[b]{0.45\linewidth}
    \centering
    \includegraphics[width=1\textwidth, ]{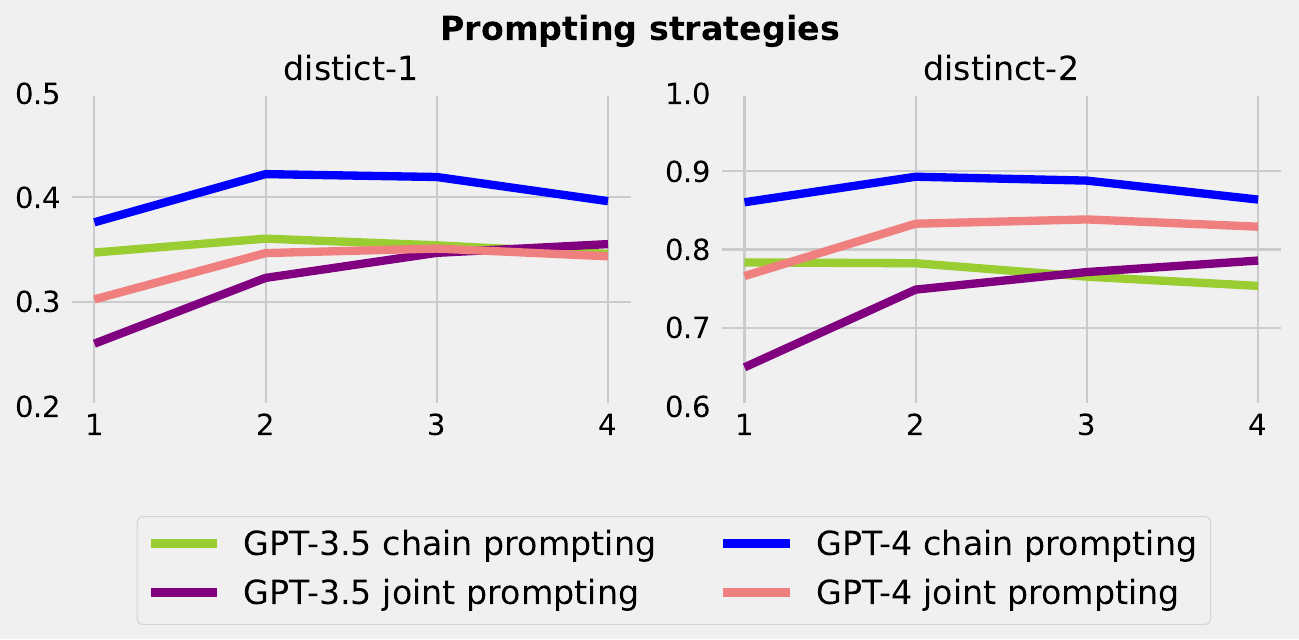}
    \caption{Dynamics of diversity for \untrainable agents over varying prompting strategies based on GPT-3.5 and GPT-4.}\label{fig:diveristy_prompting}
    \end{minipage}
    \hspace{0.5cm}
\begin{varwidth}[b]{0.45\linewidth}
    \centering
    \small
    \begin{tabular}{llll}
    \toprule
    model & strategy & distinct-1 & distinct-2 \\
    \midrule
    GPT-3.5 & chain &	0.352 &	0.771 \\
    GPT-3.5 & joint &	0.321 & 0.739 \\
    GPT-4 & chain &	0.404 &	0.876 \\
    GPT-4 & joint &	0.336 &  0.817\\
    \arrayrulecolor{black}\bottomrule
    \end{tabular}
    \captionof{table}{Diversity results in aggregative mean for \untrainable agents. \textit{Distinct-1} and \textit{distinct-2} are the percentage of distinct uni-/bi-grams.}\label{tab:diveristy_conver}
\end{varwidth}
\end{figure}
As \untrainable agents do not involve further pretraining, novelty metrics, which involve comparison to the pretraining dataset, are thus undefined. Therefore, we only study the lexical diversity of the generated poetry. Figure \ref{fig:diveristy_prompting} shows the dynamics of diversity over varying prompting strategies for agents based on GPT-3.5 and GPT-4. \newline

\textit{Do we observe an increasing trend for \untrainable agents similar to that of the \trainable agents?} Different from the trend we observe for \trainable agents, \untrainable agents exhibit a sharp increase from $t=1$ to $t=2$ with a maximum of 6.3 pp increase in \textit{distinct-1} and 10 pp in \textit{distinct-2} for nearly all experiments. GPT-3.5 under \pchain is an exception where we observe a constant decreasing trend in \textit{distinct-2}. However, the increment in lexical diversity pauses when $t>2$ where we yield slightly decreasing trends for nearly all experiments. GPT-3.5 under \pjoint is an exceptional case where the increasing trend continues mildly. We examine the effect of positive and negative learning strategies in separation in Section \ref{prompting_sep_str}. \newline

\textit{Which prompting strategy and base model perform better according to lexical diversity?} Both Figure \ref{fig:diveristy_prompting} and Table \ref{tab:diveristy_conver} indicate that GPT-4 under \pchain generates the most lexically diverse poetry compared to other settings. In general, the \pchain strategy performs better than \pjoint according to \textit{distinct-1} and \textit{distinct-2}. However, GPT-4 does not always outperform GPT-3.5 as suggested by the aggregative mean in Table \ref{tab:diveristy_conver} where GPT-3.5 under \pchain strategy delivers the second best performance according to \textit{distinct-1}. 

For \untrainable agents, our framework only benefits the generation process in a limited manner (when $t=1,2$) according to lexical diversity. Worth noting is that \untrainable agents have an overall higher percentage of unique uni-grams \textit{distinct-1} and bi-grams \textit{distinct-2} shown in Table \ref{tab:diveristy_conver}, especially for \textit{distinct-1} with below 20 pp for \trainable agents and over 40 pp for \untrainable agents. 

\subsubsection{Group divergence in semantics} 
\begin{figure}[h]
    \begin{subfigure}{0.32\textwidth}
    \centering
    \includegraphics[width=1\textwidth,height=4cm, ]{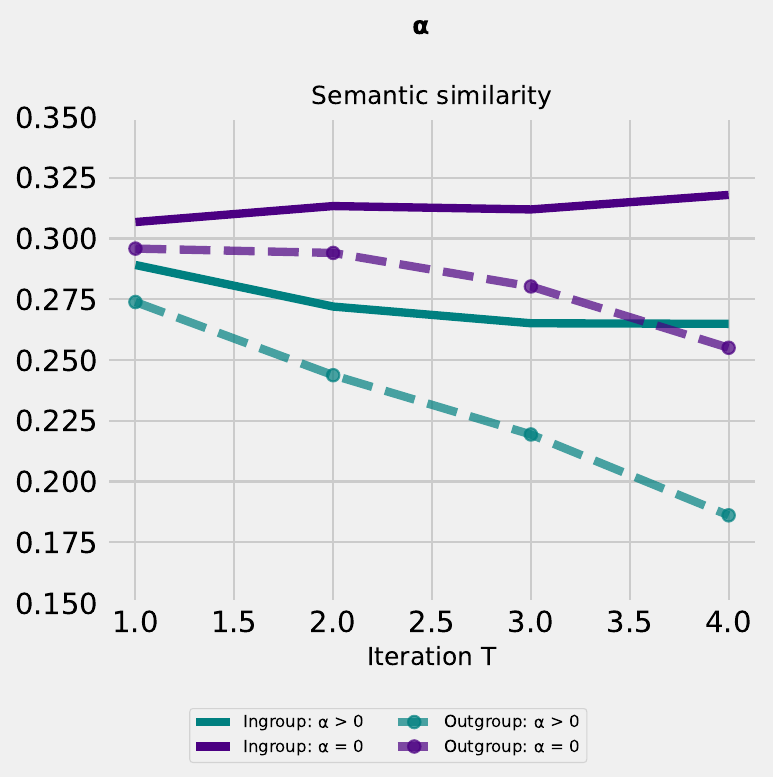}
    \caption{}\label{fig:alpha}
\end{subfigure}
\hfill
\begin{subfigure}{0.32\textwidth}
    \centering
    \includegraphics[width=1\textwidth, height=4cm, ]{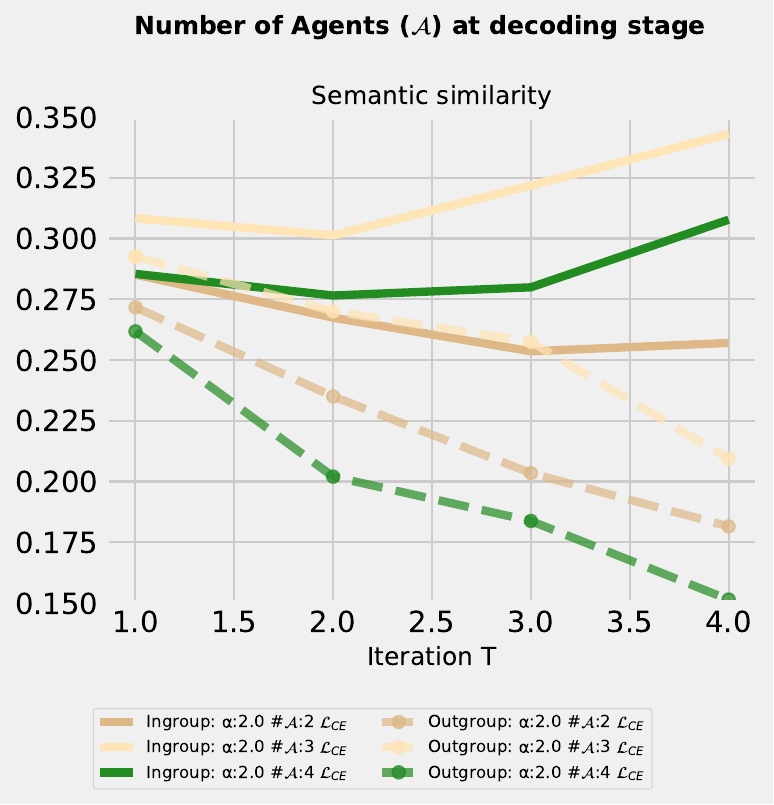}
    \caption{}\label{fig:nagent}
\end{subfigure}
\hfill
    \begin{subfigure}{0.32\textwidth}
    \centering
    \includegraphics[width=1\textwidth, height=4cm,]{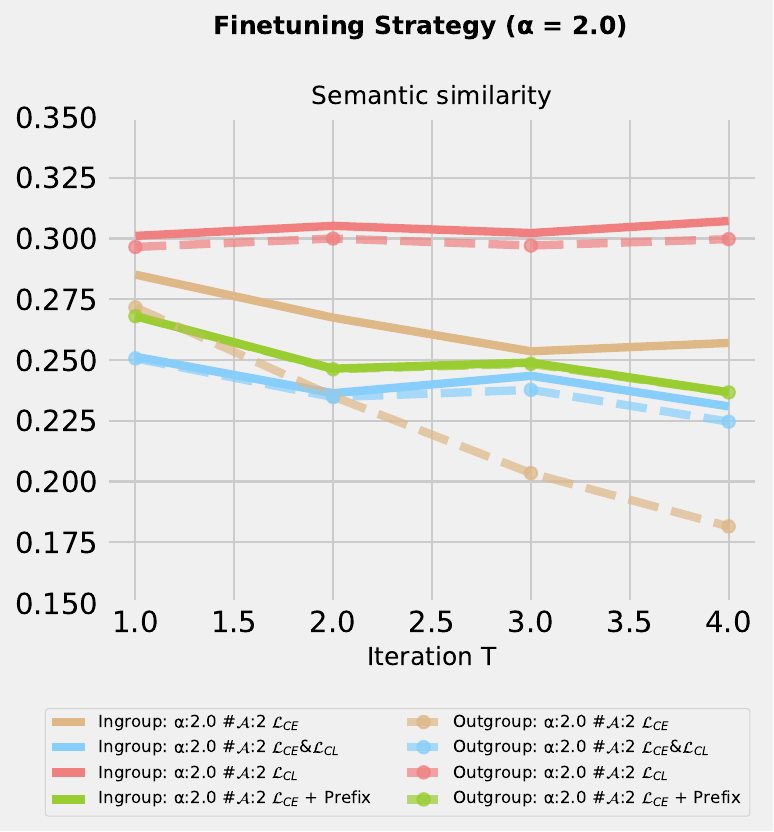}
    \caption{}\label{fig:finetuning}
\end{subfigure}
\caption{Divergence of \trainable agents measured by mean of pairwise semantic similarity over varying training parameters. (a) The effect of scaling parameter $\alpha$ in Equation (\ref{equ:decode}). (b) The effect of the number of interactive agents \#$\mathcal{A}$ during the decoding stage. (c) The effect of finetuning strategies: \lce and \lcl indicate \ce loss and contrastive loss. Prefix refers to the conditioned finetuning. The solid line and dashed line represent semantic similarity measured for \ingroup and \outgroup affiliations respectively. }\label{fig:divergence}
\end{figure}
\noindent\textbf{RQ3: how do different learning strategies affect the group dynamics of \trainable agents?}
\begin{itemize}
    \item \textit{Observable group dynamics for positive training (\ce loss) with negative decoding.} 
    Figure \ref{fig:divergence} shows the mean semantic similarity based on group affiliations for different scaling parameters $\alpha$, the number of agents $\#\mathcal{A}$ involved during the decoding stage and different finetuning strategies. The solid line represents semantic similarity measured for \ingroup agents and the dashed line for \outgroup agents.  Overall, we observe a divergence between \ingroup and \outgroup similarity for \ce loss with negative decoding under varying scaling parameters $\alpha$ and different numbers of agents $\#\mathcal{A}$. The effects of parameters vary: 1) Figure \ref{fig:alpha} exhibits the dynamics for different $\alpha$. We observe a divergence between the semantic similarity of \ingroup and \outgroup where particularly, \outgroup similarity decreases over iterations. $\alpha=0$ represents the case for `echo chambers' where only positive finetuning is considered (i.e., agents only talk to their \ingroup). For $\alpha=0$, the agents echo their own `thoughts' resulting in an overall higher level of similarity for both \ingroup and \outgroup compared to $\alpha>0$. For $\alpha>0$, we yield an 8.8 pp decrease in semantic similarity for \outgroup which is 4.7 pp greater in divergence compared to the case for $\alpha=0$ (4.1 pp in total); 2) \se{w}e observe from Figure \ref{fig:nagent} that interaction involving more agents during the decoding stage has a slightly positive influence on group divergence. $\#\mathcal{A}=4$ yields a mild \textit{increase} with 2.2 pp in \ingroup semantic similarity and an 11.0 pp decrease in \outgroup similarity (13.2 divergence in total). In contrast, $\#\mathcal{A}=2$ results in an increase with 2.8 pp for \ingroup and 9 pp decrease for \outgroup (11.8 divergence in total). Overall, $\#\mathcal{A}=4$ exhibits a lower level of similarity compared to $\#\mathcal{A}=2$.  
    \item \textit{Inseparable \ingroup and \outgroup dynamics resulting from other joint finetuning strategies.} Figure \ref{fig:finetuning} shows the outcome for different finetuning strategies involving multiple losses $\mathcal{L}$ and conditioned finetuning (i.e., Prefix). Except for the case using \lce alone as the finetuning loss (i.e., positive finetuning defined in Table \ref{tab:setup}), all other cases with joint finetuning exhibit inseparable dynamics between \ingroup and \outgroup similarity. We suspect that for contrastive learning (\lcl), a negative pair built purely based on group affiliation fails to provide enough contrastivity considering that we initiate the agents in a random manner. Such random initialization may affect the results for conditioned finetuning as well. 
\end{itemize}

\begin{figure}
    \centering
    \includegraphics[width=0.35\linewidth, ]{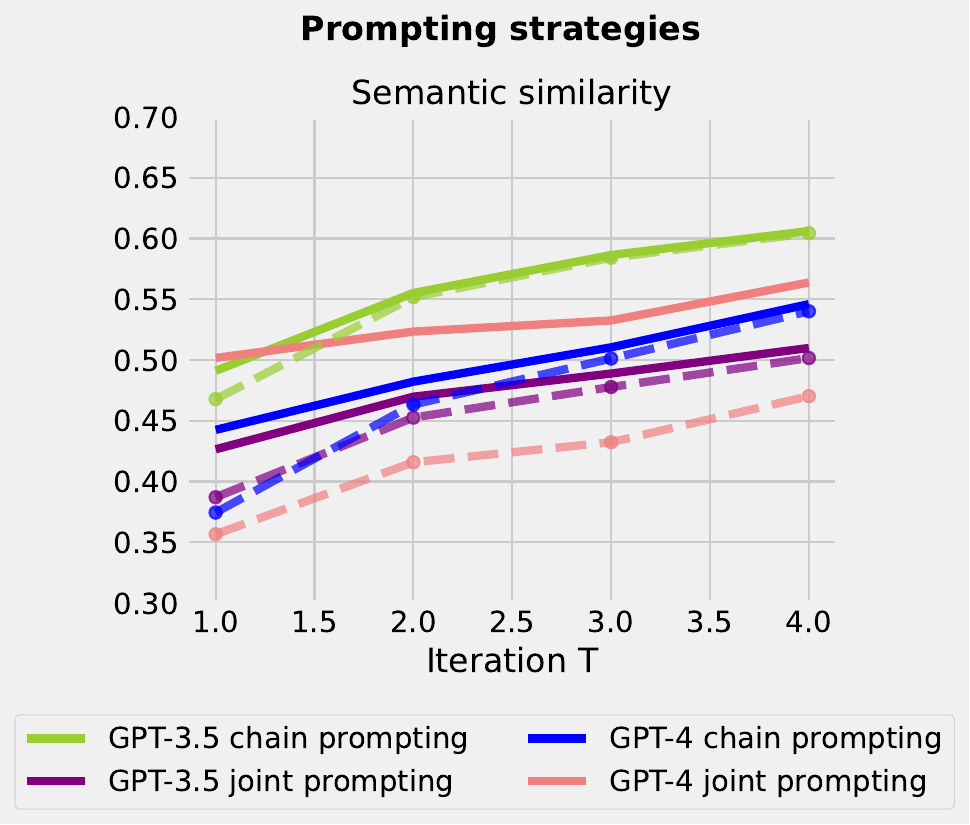}
    \caption{Group divergence of \untrainable agents measured by the mean of pairwise semantic similarity over varying prompting strategies and base model. The solid line and dashed line represent semantic similarity measured for \ingroup and \outgroup affiliations respectively.}\label{fig:divergence_prompting}
\end{figure}

\noindent\textbf{RQ4: how do different prompting strategies affect the group dynamics of \untrainable agents?} \newline

\textit{Undesirable increasing semantic similarity from \outgroup agents} Figure \ref{fig:divergence_prompting} shows the group divergence of \untrainable agents measured by the mean pairwise semantic similarity over varying prompting strategies and base models. The solid line represents the semantic similarity measured for
\ingroup agents and the dashed line for \outgroup agents. We observe an increasing similarity for both \ingroup and \outgroup agents where the greatest group divergence is observed at $t=1$. Over time, the agents tend to generate semantically similar poetry for both GPT-3 and GPT-4. Moreover, we also notice that \untrainable agents generate poetry of homogeneous styles over time which coincides the finding of \citet{sawicki2023bits}. We discuss this point in more detail in Section \ref{sec:qualitative}. 

\subsection{Qualitative evaluation}\label{sec:qualitative}
\begin{table}
    \small 
    \begin{tabular}{p{.1cm}p{6.5cm}p{6.5cm}}
    \toprule
        t & group A & group B  \\
        \midrule
        \multicolumn{3}{c}{\textit{\trainable agents ($\alpha =2,  \#\mathcal{A}=2$, \lce)}} \\
        0 &  \thead{That will be and the world shall give place to one, \\
        whom death has never forgotten. \\
Here \textit{thy} dearest child is alone: \\A sad birth \textit{thou} \textit{seekest} still more mourn}
        & \thead{\textit{Thou}, the child of Israel \\
        - \textit{thou} who \textit{dost} keep the Lord forever; \\
        And every day we mourn \textit{thy} loss.} \\
        1 & \thead{vultures flying;
 their cries in one another's \textcolor{gray}{midst}: \\
And when \textit{thou} \textit{seeest} a poor child,\\
\textcolor{red}{Who'd the hand that's to feed him \textcolor{gray}{must}}?}
        & \thead{The children of the world have been lulled to sleep. \\ In sorrow's night their eyes fail to see \\
        - no tears are shed, but only dreams; \\ 
And they will say that nothing is more beautiful}\\
        4 & \thead{
Pretend, I know not where: \\
But this is what has made me love \textit{thee} so;\\
For \textit{thou art} a child of my breast,\\
The affection I have felt for her in youth's day.} & \thead{reputations will be lost; \\
 lives of men,  women and children shall be gone. \\
But the memory of my life on \textcolor{gray}{earth} \\
Will grow in faithfulness like a tree-top  \\
That grows new leaves that come to rest \\
Where \textcolor{red}{its} roots \textcolor{gray}{once}} \\ 
\multicolumn{3}{c}{\textit{\untrainable agents (GPT-3.5 chain-prompting)}} \\
0 & \thead{Tiny hands and curious \textbf{hearts},\\
Youthful souls, creating \textbf{art}, \\
Innocent laughter fills the \textbf{air},\\
Children's world, free from \textbf{care}, \\
Growing up, but never \textbf{old}, \\
Their stories waiting to be \textbf{told}.}& 
\thead{Youth's fleeting flame burns bright and \textbf{bold}, \\
Innocence wrapped in dreams \textbf{untold}. \\
Carefree laughter, hearts pure and free, \\
Embracing life's wonder, wild and carelessly. \\
Oh, to be young, to feel so \textbf{alive}, \\
In a world where anything can \textbf{thrive}. }  \\
1 &  \thead{Tears, like a river, \textbf{flowed} \\ Father's decree, "Stay away, my \textbf{child}" \\ Through the wild woods he \textbf{roamed} \\ The wind, his sole confidant \\ He races, longing for home, heart \textbf{mild}} & 
\thead{ Arise, oh heart! \\ Unveil thy pure essence, \\as we danced in our vibrant youth.}\\
4 &\thead{In the moon's gentle \textbf{light}, \\
I dance with shadows in the \textbf{night}, \\
With a heart filled with a \textbf{melody}, \\
I roam through nature's \textbf{symphony}.\\}& 
\thead{Lost in the shadows of a world unknown, \\
   Drifting through the silence all alone, \\
   Seeking solace in the whispers of the \textbf{night}, \\
   Longing for a glimpse of dawn's soft \textbf{light}.\\}\\
\bottomrule \\
    \end{tabular}
    \caption{Generated poetry based on group affiliation from \trainable and \untrainable agents at different iteration $t$. For \untrainable agents, the baseline at $t=0$ is generated using one simple prompt (`Please generate a poem about children or youth''). We highlight words containing historical components in \textit{italics}. Words that rhyme are highlighted in \textbf{bold}. Words that mildly rhyme are colored in \textcolor{gray}{gray}. Grammatical errors are marked in \textcolor{red}{red}.}
    \label{tab:qualitative}
\end{table}
Table \ref{tab:qualitative} contains examples of generated poetry from \trainable agents under positive finetuning and negative decoding strategy (i.e., $\alpha =2, \#\mathcal{A}=2$, \lce) and from \untrainable agents using GPT-3.5 under the \pchain strategy. We select examples composed with similar themes, i.e., child or youth. Poems generated at $t=0$ are considered the baselines for both  \trainable and \untrainable frameworks. 

For \trainable agents, at $t=0$ the generated poems often contain historical spellings (e.g., `thy' and `thou') and historical morphology terms (e.g., `seekest' and `dost'). Apart from the semantic divergence discussed in the previous section, we observe a divergence in word choices over time. For example, we study the poems generated by \trainable agents with settings $\alpha =2,  \#\mathcal{A}=2$, \lce where we calculate the percentage of poems that contain historical spellings and historical morphology terms. We find that over 26\% of poems generated by agents in group A contain historical spellings or morphology terms compared to only 10\% of poems by agents from group B. Moreover, the percentage of poems with historical languages for group A is stable at a level of 26\% over iterations while for group B, the percentage steadily decreases over time by nearly 6 pp. The word frequency of poems from group A and group B also suggests such divergence. For example, words such as ``mind, thy, thee, nature, art, power, happy, hath, young, pleasant, friend, ...'' are more frequent in group A and less frequent in \se{g}roup B while in group B the more frequent words are ``god, lord, light, sun, sky, sea, land, soul, children, dream, ...''. For \untrainable agents, in contrast, we observe more diverse lexicons for both groups compared to \trainable agents but the two groups in \untrainable setting hardly diverge in terms of vocabulary or topics when $t>1$. This is also suggested by Figure \ref{fig:divergence_prompting}. Moreover, \untrainable agents tend to generate poems of homogenous styles over time. As shown in Table \ref{tab:qualitative}, poems generated from \untrainable agents excessively focus on rhymes which makes the generated poetry merely superficially human-like. This also suggests a poor understanding of poetry for GPT-3.5 and GPT-4 in a zero-shot setting. Even though GPT-3.5 and GPT-4 can adopt historical texts well \cite{zhang2023cross}, they never pick up the historical expressions from the initial poetry as the \trainable agents do. Apart from the `obsession' for rhyming, GPT-3.5 and GPT-4 also tend to generate poems using similar beginning phrases such as ``Beneath/Under XXX, In the XXX, Lost in XXX'', especially when $t>1$. The generated poems from GPT-3.5 and GPT-4 contain fewer grammatical errors than \trainable agents, though \trainable agents generate poems of more diverse styles and topics in comparison.

\section{Discussion and analysis}
\label{section:analysis}

\subsection{How stable are the simulation results?} \label{sec:stability}   
Due to resource constraints, we do not execute multiple simulations for all experiment settings. Instead, we study the stability of our experiment\se{s} using two experiment settings for \trainable agents, i.e., $\alpha =0$ and $\alpha=2$\se{,} and one experiment setting for \untrainable agents, i.e., GPT-3,5 \pchain. We rerun the experiments three times under the same parameters (or prompt templates for \untrainable agents) and initialization. We then yield three sets of statistics and calculate the standard deviation as our stability measure. We study the stability from two perspectives: 1) stability of the aggregative mean and 2) dynamic stability. \newline

\begin{table}\centering
\small
\begin{tabular}{p{2.3cm}p{2.3cm}p{1cm}p{1cm}p{1cm}p{1cm}}
\toprule
&model \& setting &distinct-1 &distinct-2 &novelty-1 &novelty-2 \\
\arrayrulecolor{black! 100}\midrule
\multirow{2}{*}{\trainable} &$\alpha = 0$ &0.001 (.120)&0.002 (.664) &0.001 (.034) &0.003 (.136) \\
&$\alpha = 2$ &0.003 (.164) &0.004 (.709) &0.004 (.095) &0.005 (.238) \\
\arrayrulecolor{black! 10}\midrule
\untrainable & \makecell{GPT-3.5 \\ \pchain} &0.007 (.322) &0.007 (.755) &- &- \\
\arrayrulecolor{black! 100}\bottomrule \\
\end{tabular}
\caption{Stability of three simulation results measured by standard deviation. The mean values of all three simulations are reported in brackets.}\label{tab:stab_agg}
\end{table}

\noindent\textbf{Stability of the aggregative mean.} 
The stability results of the aggregative mean are shown in Table \ref{tab:stab_agg}. We observe a low level of variation with less than 0.7 pp for both \trainable and \untrainable agents. \newline

\noindent\textbf{Dynamic stability.} 
\begin{table}
\centering
\small
\begin{tabular}{p{2.5cm}p{2.5cm}p{0.5cm}p{1cm}p{1cm}p{1cm}p{1cm}}
\toprule
&\textbf{model \& setting} &\textbf{t} &distinct-1 &distinct-2 &novelty-1 &novelty-2 \\\midrule
\multirow{8}{*}{\trainable} &\multirow{4}{*}{$\alpha =0$} &1  &0.001 (.125)&0.003 (.684) &0.001 (.036) &0.003 (.129) \\
& &2 &0.001 (.120) &0.002 (.666)&0.002 (.032) &0.001 (.128) \\
& &3 &\textbf{0.005} (.118) &\textbf{0.006} (.660) &0.001 (.034) &0.004 (.136) \\
& &4 & 0.001 (.116) &0.005 (.647) &\textbf{0.005} (.034) &\textbf{0.005} (.151) \\
&\multirow{4}{*}{$\alpha=2$} &1 &0.002 (.147) &0.005 (.699) &0.003 (.065) &0.001 (.183)\\
& &2 &0.001 (.162)&0.002 (.708) &0.005 (.081) &0.005 (.217) \\
& &3 &\textbf{0.007} (.175) &\textbf{0.008} (.719) &\textbf{0.007} (.106) &\textbf{0.010} (.255) \\
& &4 &0.001 (.172) &0.003 (.708)&0.005 (.126)&0.009 (.298)\\
\arrayrulecolor{black! 10}\midrule
\multirow{4}{*}{\untrainable} &\multirow{4}{*}{\makecell{GPT-3.5 \\ chain-prompting}} &1 &\textbf{0.019} (.338) &0.006 (.792)&-&- \\
& &2 &0.011 (.328) &0.002 (.763) &-&- \\
& &3 &0.017 (.317) &\textbf{0.007} (.740)  &-&- \\
& &4 &0.008 (.304) &0.004 (.724)  &-&- \\
\arrayrulecolor{black! 100}\bottomrule \\
\end{tabular}
\caption{Dynamic stability of three simulation results measured by standard deviation. The highest value in each experimental setting is highlighted in \textbf{bold}. The mean values of all three simulations are reported in brackets. }\label{tab:stab_dyn}
\end{table}
The stability results for our dynamic statistics are shown in Table \ref{tab:stab_dyn}. We highlight the highest value in each setting in \textbf{bold}. For \trainable agents, the results show a low variation with a maximum of 1 pp. Results at iterations 3 and 4 show a slightly higher variation for all four measures than the results at $t=1,2$. 
In contrast, for \untrainable agents, we observe a greater level of variation with the highest standard deviation to the level of 1.9 pp. Specifically, the results for \textit{distinct-1} exhibit more instability than other measures. This may be caused by the more diverse lexicons from \untrainable agents.  \newline

Overall, our simulations indicate high stability over the statistics, especially for \trainable agents. The \untrainable agents are slightly more unstable (with a variation up to 1.9 pp) in comparison to \trainable agents (with a maximum variation of 1 pp and 80\% of the variation under 0.5 pp). \newline 

\subsection{The effect of different learning strategies and heterogeneous models for \untrainable agents}\label{prompting_sep_str} 
\begin{table}[]
    \centering
    \small
    \begin{tabular}{llll}
    \toprule
    model & strategy & distinct-1 & distinct-2 \\
    \midrule
    GPT-4 & positive  &	0.286 &	0.598 \\
    GPT-4 & negative &	0.313 &  0.653\\
    GPT-4 & joint (positive + negatives) &	0.336 &  0.817\\
    \arrayrulecolor{black}\bottomrule \\
    \end{tabular}
    \captionof{table}{Diversity results in aggregative mean for \untrainable agents under different learning strategies. \textit{Distinct-1} and \textit{distinct-2} are the percentage of distinct uni-/bi-grams.}\label{tab:analysis_sep}
\end{table}

\noindent\textbf{Non-cooperative environments boost diversity. } To examine the effect of learning strategies for \untrainable agents, we utilize the same experimental parameters as in Section \ref{sec:setup} and conduct generation under positive-only, negative-only, and joint learning strategies (\pjoint), respectively. As shown in Table \ref{tab:analysis_sep}, the joint learning strategy, which employs both positive and negative steps, is the most effective in terms of the diversity of the generated poetry, yielding a 5.0 pp increase in \textit{distinct-1} and over a 20 pp increase in \textit{distinct-2} compared to the positive-only strategy. Moreover, the negative-only strategy enhances diversity compared to the positive-only strategy, but to a lesser extent than the joint approach. \\

\begin{table}[]
    \centering
    \small
    \begin{tabular}{lll}
    \toprule
    model  & distinct-1 & distinct-2 \\
    \midrule
    GPT-4  &	0.336 &  0.817 \\
    GPT-4 + GPT-3.5  &	0.413 &  0.814\\
    GPT-4 + GPT-3.5 + LlaMa3-7b  & 0.511 & 0.887 \\
    \arrayrulecolor{black}\bottomrule \\
    \end{tabular}
    \captionof{table}{Diversity results in aggregative mean for \untrainable agents with heterogeneous models.}\label{tab:analysis_hetero}
\end{table}

\noindent\textbf{Heterogeneous models can boost the diversity of the system.} To test the effect of using non-\zr{homogeneous} agents, we use a combination of various models to conduct experiments using \pjoint defined in Section \ref{sec:setup}. As shown in Table \ref{tab:analysis_hetero}, when GPT-4 is combined with GPT-3.5, the \textit{distinct-1} score increases by 7.7 pp to 0.413, while \textit{distinct-2} slightly decreases by 0.3 pp to 0.814. Incorporating LlaMa3-7b along with GPT-4 and GPT-3.5 further enhances the diversity, with \textit{distinct-1} increasing by an additional 9.8 pp to 0.511, and \textit{distinct-2} increasing by 7.3 pp to 0.887. This demonstrates the potential benefits of employing a more diverse ensemble of models.

\subsection{Can different initializations lead to group-based behaviors for \untrainable agents?} 
As discussed in Section \ref{sec:results}, the framework built with \untrainable agents does not exhibit any group-based behavior as expected. Considering that we initialize the agents with random samples drawn from the \quatrain corpus, we suspect this may cause a high resemblance among the initialized poems. To examine whether an initialization with poems of more contrastive forms can produce group-based behavior, we conduct an experiment using GPT-3.5 under \pchain strategy where we initialize group A with poems written by \textit{Edgar Allan Poe} and group B with poems written by school children under 12 years old \cite{hipson-mohammad-2020-poki}. An example poem from Edgar Allan Poe is \textit{``From the lightning in the sky, As it passed me flying by, From the thunder and the storm, And the cloud that took the form.''} and an example poem from a school child is  \textit{``Roses are red, violets are blue. I love the zoo. do you?''} We implement the same process and compute the statistics. Slightly surprisingly, we observe a very similar trend for both diversity and semantic divergence to that of random initialized \untrainable agents as shown in Section \ref{sec:results}. In terms of diversity, we notice an increase of 2 pp at iteration $t=1$ and then a decreasing trend for both \textit{distinct-1} and \textit{distinct-2}. Qualitatively, at iteration $t=1$, we obtain poems from group B such as  \textit{``As the sun rose, a butterfly landed softly on my hand, whispering secrets of the garden with each flutter of its delicate wings.''}, which resembles the tone of a child and the imagery of a child playing in the garden. However, as $t>1$, we yield similar homogeneous poems to the case in Table \ref{tab:qualitative}. An example poem from group B at $t=4$ is \textit{``Beneath the starlit sky, a solitary figure stands,
A soft whisper of wind caresses the quiet lands.
Burdened with untold sorrows in the night so still,
`I am but a fleeting shadow, lost in time's skill.' ''}. The results again suggest that GPT-3.5 (also GPT-4) tends to ignore the prompts (i.e., in our case their personas) and rely more on its pretraining knowledge. This observation coincides with the work from \citet{chuang2023simulating,tirumala2022memorization} that larger models suffer more from memorization.       
\section{Concluding remarks}\label{sec:conclusion}
In this paper, we introduce an LLM-based multi-agent framework that incorporates not only cooperative interaction but also non-cooperative environments. We experiment with $M=4$ \trainable agents trained on GPT-2 and \untrainable agents employing GPT-3.5 and GPT-4. Our evaluation with 96k generated poems shows that for \trainable agents: 1) non-cooperative environments encourage diversity and novelty over iteration measured by distinct and novel n-grams; 2) \trainable agents demonstrate group divergence in terms of lexicons, styles and semantics in accordance to the predefined group affiliation. Our results also indicate that for \untrainable agents: 3) the generated poetry contains very few grammatical errors with a more diverse lexicon; 4) the \untrainable framework benefits from non-cooperative environments and heterogeneous model in terms of aggregated diversity; 5) dynamically, \untrainable framework barely improves lexical diversity after the first iteration and unlike \trainable agents, \untrainable agents do not show group-based divergence as expected; 6) \untrainable agents are prone to generating poetry of more homogeneous styles over time, presumably suggesting the memorization problems of LLMs. 

Nowadays, more researchers have raised concerns that the use of LLMs may lead to homogeneity and uniformity of human language and knowledge \cite{kuteeva2024diversity}. Empirical evidence also suggests that LLMs under the current training paradigm such as RLHF (i.e., reinforcement learning from
human feedback) produce less diverse text \cite{kirk2023understanding, 2023evalDiv}. In this context, we believe a training paradigm shift towards a more human-like machine-learning process, particularly for creative tasks such as poetry generation, is thus necessary and meaningful. As suggested by our experiments, a more human-like (network-structured) social learning process that emphasizes non-cooperative interaction can bring in more diversity and novelty. Our results also show promise for mitigating the issues of data degeneration caused by the `self-consuming' loop during modeling \cite{wang2022self}.  

Future work can improve on several points. For \trainable agents, enhancing inference efficiency using techniques such as speculative sampling would benefit the scaling of the framework \cite{dekoninck2023controlled} and thus boost diversity and novelty to a greater level. For \untrainable agents, involving more complex reasoning methods such as tree-of-thought \cite{yao2024tree} into the prompting might be helpful. Extending the current framework to include an interactive combination of both \trainable and \untrainable agents might be interesting to explore in which a diverse network of LLMs might bring additional generation diversity to the system.

\bibliographystyle{plainnat}
\bibliography{references}  

\begin{thebibliography}{92}
\providecommand{\natexlab}[1]{#1}
\providecommand{\url}[1]{\texttt{#1}}
\expandafter\ifx\csname urlstyle\endcsname\relax
  \providecommand{\doi}[1]{doi: #1}\else
  \providecommand{\doi}{doi: \begingroup \urlstyle{rm}\Url}\fi

\bibitem[Amirkhani and Barshooi(2022)]{amirkhani2022consensus}
Abdollah Amirkhani and Amir~Hossein Barshooi.
\newblock Consensus in multi-agent systems: a review.
\newblock \emph{Artificial Intelligence Review}, 55\penalty0 (5):\penalty0 3897--3935, 2022.

\bibitem[Baker(1972)]{baker1972hemingway}
Carlos Baker.
\newblock \emph{Hemingway, the writer as artist}.
\newblock Princeton University Press, 1972.

\bibitem[Belouadi and Eger(2023)]{belouadi-eger-2023-bygpt5}
Jonas Belouadi and Steffen Eger.
\newblock {B}y{GPT}5: End-to-end style-conditioned poetry generation with token-free language models.
\newblock In Anna Rogers, Jordan Boyd-Graber, and Naoaki Okazaki, editors, \emph{Proceedings of the 61st Annual Meeting of the Association for Computational Linguistics (Volume 1: Long Papers)}, pages 7364--7381, Toronto, Canada, July 2023. Association for Computational Linguistics.
\newblock \doi{10.18653/v1/2023.acl-long.406}.
\newblock URL \url{https://aclanthology.org/2023.acl-long.406}.

\bibitem[Bena and Kalita(2019)]{bena2019introducing}
Brendan Bena and Jugal Kalita.
\newblock Introducing aspects of creativity in automatic poetry generation.
\newblock In \emph{Proceedings of the 16th International Conference on Natural Language Processing}, pages 26--35, 2019.

\bibitem[Biesialska et~al.(2020)Biesialska, Biesialska, and Costa-juss{\`a}]{biesialska2020continual}
Magdalena Biesialska, Katarzyna Biesialska, and Marta~R Costa-juss{\`a}.
\newblock Continual lifelong learning in natural language processing: A survey.
\newblock In \emph{Proceedings of the 28th International Conference on Computational Linguistics}, pages 6523--6541, 2020.

\bibitem[Brinkmann et~al.(2023)Brinkmann, Baumann, Bonnefon, Derex, M{\"u}ller, Nussberger, Czaplicka, Acerbi, Griffiths, Henrich, et~al.]{brinkmann2023machine}
Levin Brinkmann, Fabian Baumann, Jean-Fran{\c{c}}ois Bonnefon, Maxime Derex, Thomas~F M{\"u}ller, Anne-Marie Nussberger, Agnieszka Czaplicka, Alberto Acerbi, Thomas~L Griffiths, Joseph Henrich, et~al.
\newblock Machine culture.
\newblock \emph{Nature Human Behaviour}, 7\penalty0 (11):\penalty0 1855--1868, 2023.

\bibitem[Chakrabarty et~al.(2021)Chakrabarty, Saakyan, and Muresan]{chakrabarty2021don}
Tuhin Chakrabarty, Arkadiy Saakyan, and Smaranda Muresan.
\newblock Don’t go far off: An empirical study on neural poetry translation.
\newblock In \emph{Proceedings of the 2021 Conference on Empirical Methods in Natural Language Processing}, pages 7253--7265, 2021.

\bibitem[Chakrabarty et~al.(2022)Chakrabarty, Padmakumar, and He]{chakrabarty-etal-2022-help}
Tuhin Chakrabarty, Vishakh Padmakumar, and He~He.
\newblock Help me write a poem: Instruction tuning as a vehicle for collaborative poetry writing.
\newblock In Yoav Goldberg, Zornitsa Kozareva, and Yue Zhang, editors, \emph{Proceedings of the 2022 Conference on Empirical Methods in Natural Language Processing}, pages 6848--6863, Abu Dhabi, United Arab Emirates, December 2022. Association for Computational Linguistics.
\newblock \doi{10.18653/v1/2022.emnlp-main.460}.
\newblock URL \url{https://aclanthology.org/2022.emnlp-main.460}.

\bibitem[Chakrabarty et~al.(2023)Chakrabarty, Padmakumar, He, and Peng]{chakrabarty2023creative}
Tuhin Chakrabarty, Vishakh Padmakumar, He~He, and Nanyun Peng.
\newblock Creative natural language generation.
\newblock In \emph{Proceedings of the 2023 Conference on Empirical Methods in Natural Language Processing: Tutorial Abstracts}, pages 34--40, 2023.

\bibitem[Chan et~al.(2023)Chan, Chen, Su, Yu, Xue, Zhang, Fu, and Liu]{chan2023chateval}
Chi-Min Chan, Weize Chen, Yusheng Su, Jianxuan Yu, Wei Xue, Shanghang Zhang, Jie Fu, and Zhiyuan Liu.
\newblock Chateval: Towards better llm-based evaluators through multi-agent debate.
\newblock \emph{arXiv preprint arXiv:2308.07201}, 2023.

\bibitem[Chen et~al.(2024)Chen, Gr{\"o}ner, Zarrie{\ss}, and Eger]{2023evalDiv}
Yanran Chen, Hannes Gr{\"o}ner, Sina Zarrie{\ss}, and Steffen Eger.
\newblock Evaluating diversity in automatic poetry generation.
\newblock \emph{arXiv preprint arXiv:2406.15267}, 2024.

\bibitem[Chronopoulou et~al.(2023)Chronopoulou, Peters, Fraser, and Dodge]{chronopoulou2023adaptersoup}
Alexandra Chronopoulou, Matthew~E Peters, Alexander Fraser, and Jesse Dodge.
\newblock Adaptersoup: Weight averaging to improve generalization of pretrained language models.
\newblock In \emph{Findings of the Association for Computational Linguistics: EACL 2023}, pages 2054--2063, 2023.

\bibitem[Chuang et~al.(2023)Chuang, Goyal, Harlalka, Suresh, Hawkins, Yang, Shah, Hu, and Rogers]{chuang2023simulating}
Yun-Shiuan Chuang, Agam Goyal, Nikunj Harlalka, Siddharth Suresh, Robert Hawkins, Sijia Yang, Dhavan Shah, Junjie Hu, and Timothy~T Rogers.
\newblock Simulating opinion dynamics with networks of llm-based agents.
\newblock \emph{arXiv preprint arXiv:2311.09618}, 2023.

\bibitem[Chuang et~al.(2024)Chuang, Harlalka, Suresh, Goyal, Hawkins, Yang, Shah, Hu, and Rogers]{chuang2024wisdom}
Yun-Shiuan Chuang, Nikunj Harlalka, Siddharth Suresh, Agam Goyal, Robert~D Hawkins, Sijia Yang, Dhavan~V Shah, Junjie Hu, and Timothy~T Rogers.
\newblock The wisdom of partisan crowds: Comparing collective intelligence in humans and llm-based agents.
\newblock In \emph{ICLR 2024 Workshop on Large Language Model (LLM) Agents}, 2024.

\bibitem[Dathathri et~al.(2019)Dathathri, Madotto, Lan, Hung, Frank, Molino, Yosinski, and Liu]{dathathri2019plug}
Sumanth Dathathri, Andrea Madotto, Janice Lan, Jane Hung, Eric Frank, Piero Molino, Jason Yosinski, and Rosanne Liu.
\newblock Plug and play language models: A simple approach to controlled text generation.
\newblock In \emph{International Conference on Learning Representations}, 2019.

\bibitem[Dekoninck et~al.(2023)Dekoninck, Fischer, Beurer-Kellner, and Vechev]{dekoninck2023controlled}
Jasper Dekoninck, Marc Fischer, Luca Beurer-Kellner, and Martin Vechev.
\newblock Controlled text generation via language model arithmetic.
\newblock In \emph{The Twelfth International Conference on Learning Representations}, 2023.

\bibitem[Dinan et~al.(2020)Dinan, Fan, Williams, Urbanek, Kiela, and Weston]{dinan2020queens}
Emily Dinan, Angela Fan, Adina Williams, Jack Urbanek, Douwe Kiela, and Jason Weston.
\newblock Queens are powerful too: Mitigating gender bias in dialogue generation.
\newblock In \emph{Proceedings of the 2020 Conference on Empirical Methods in Natural Language Processing (EMNLP)}, pages 8173--8188, 2020.

\bibitem[Du et~al.(2023)Du, Li, Torralba, Tenenbaum, and Mordatch]{du2023improving}
Yilun Du, Shuang Li, Antonio Torralba, Joshua~B Tenenbaum, and Igor Mordatch.
\newblock Improving factuality and reasoning in language models through multiagent debate.
\newblock \emph{arXiv preprint arXiv:2305.14325}, 2023.

\bibitem[Eger(2016)]{eger2016opinion}
Steffen Eger.
\newblock Opinion dynamics and wisdom under out-group discrimination.
\newblock \emph{Mathematical Social Sciences}, 80:\penalty0 97--107, 2016.

\bibitem[Elgammal et~al.(2017)Elgammal, Liu, Elhoseiny, and Mazzone]{elgammal2017can}
Ahmed Elgammal, Bingchen Liu, Mohamed Elhoseiny, and Marian Mazzone.
\newblock Can: Creative adversarial networks generating “art” by learning about styles and deviating from style norms.
\newblock In \emph{8th International Conference on Computational Creativity, ICCC 2017}. Georgia Institute of Technology, 2017.

\bibitem[Firdaus et~al.(2022)Firdaus, Chauhan, Ekbal, and Bhattacharyya]{emosen}
Mauajama Firdaus, Hardik Chauhan, Asif Ekbal, and Pushpak Bhattacharyya.
\newblock Emosen: Generating sentiment and emotion controlled responses in a multimodal dialogue system.
\newblock \emph{IEEE Transactions on Affective Computing}, 13\penalty0 (3):\penalty0 1555--1566, 2022.
\newblock \doi{10.1109/TAFFC.2020.3015491}.

\bibitem[Fu et~al.(2022)Fu, Peng, Sabharwal, Clark, and Khot]{fu2022complexity}
Yao Fu, Hao Peng, Ashish Sabharwal, Peter Clark, and Tushar Khot.
\newblock Complexity-based prompting for multi-step reasoning.
\newblock In \emph{The Eleventh International Conference on Learning Representations}, 2022.

\bibitem[Gao et~al.(2023)Gao, Lan, Lu, Mao, Piao, Wang, Jin, and Li]{gao2023s}
Chen Gao, Xiaochong Lan, Zhihong Lu, Jinzhu Mao, Jinghua Piao, Huandong Wang, Depeng Jin, and Yong Li.
\newblock S$^3$: Social-network simulation system with large language model-empowered agents.
\newblock \emph{arXiv preprint arXiv:2307.14984}, 2023.

\bibitem[Gao et~al.(2021)Gao, Yao, and Chen]{gao2021simcse}
Tianyu Gao, Xingcheng Yao, and Danqi Chen.
\newblock Simcse: Simple contrastive learning of sentence embeddings.
\newblock \emph{arXiv preprint arXiv:2104.08821}, 2021.

\bibitem[Gautier et~al.(2022)Gautier, Stephens, Lacerda, Hawes, and Wooldridge]{gautier2022negotiated}
Anna Gautier, Alex Stephens, Bruno Lacerda, Nick Hawes, and Michael Wooldridge.
\newblock Negotiated path planning for non-cooperative multi-robot systems.
\newblock 2022.

\bibitem[Ghazvininejad et~al.(2017)Ghazvininejad, Shi, Priyadarshi, and Knight]{ghazvininejad-etal-2017-hafez}
Marjan Ghazvininejad, Xing Shi, Jay Priyadarshi, and Kevin Knight.
\newblock {H}afez: an interactive poetry generation system.
\newblock In Mohit Bansal and Heng Ji, editors, \emph{Proceedings of {ACL} 2017, System Demonstrations}, pages 43--48, Vancouver, Canada, July 2017. Association for Computational Linguistics.
\newblock URL \url{https://aclanthology.org/P17-4008}.

\bibitem[Greene et~al.(2010)Greene, Bodrumlu, and Knight]{greene-etal-2010-automatic}
Erica Greene, Tugba Bodrumlu, and Kevin Knight.
\newblock Automatic analysis of rhythmic poetry with applications to generation and translation.
\newblock In Hang Li and Llu{\'\i}s M{\`a}rquez, editors, \emph{Proceedings of the 2010 Conference on Empirical Methods in Natural Language Processing}, pages 524--533, Cambridge, MA, October 2010. Association for Computational Linguistics.
\newblock URL \url{https://aclanthology.org/D10-1051}.

\bibitem[Hamilton(2023)]{hamilton2023blind}
Sil Hamilton.
\newblock Blind judgement: Agent-based supreme court modelling with gpt.
\newblock In \emph{The AAAI-23 Workshop on Creative AI Across Modalities}, 2023.

\bibitem[Hipson and Mohammad(2020)]{hipson-mohammad-2020-poki}
Will Hipson and Saif~M. Mohammad.
\newblock {P}o{K}i: A large dataset of poems by children.
\newblock In Nicoletta Calzolari, Fr{\'e}d{\'e}ric B{\'e}chet, Philippe Blache, Khalid Choukri, Christopher Cieri, Thierry Declerck, Sara Goggi, Hitoshi Isahara, Bente Maegaard, Joseph Mariani, H{\'e}l{\`e}ne Mazo, Asuncion Moreno, Jan Odijk, and Stelios Piperidis, editors, \emph{Proceedings of the Twelfth Language Resources and Evaluation Conference}, pages 1578--1589, Marseille, France, May 2020. European Language Resources Association.
\newblock ISBN 979-10-95546-34-4.
\newblock URL \url{https://aclanthology.org/2020.lrec-1.196}.

\bibitem[Janaway(2002)]{janaway2002schopenhauer}
Christopher Janaway.
\newblock \emph{Schopenhauer: A very short introduction}.
\newblock OUP Oxford, 2002.

\bibitem[Jarvis(2012)]{jarvis2012towards}
Peter Jarvis.
\newblock \emph{Towards a comprehensive theory of human learning}.
\newblock Routledge, 2012.

\bibitem[Jiang et~al.(2024)Jiang, Sablayrolles, Roux, Mensch, Savary, Bamford, Chaplot, Casas, Hanna, Bressand, et~al.]{jiang2024mixtral}
Albert~Q Jiang, Alexandre Sablayrolles, Antoine Roux, Arthur Mensch, Blanche Savary, Chris Bamford, Devendra~Singh Chaplot, Diego de~las Casas, Emma~Bou Hanna, Florian Bressand, et~al.
\newblock Mixtral of experts.
\newblock \emph{arXiv preprint arXiv:2401.04088}, 2024.

\bibitem[Jiang et~al.(2023)Jiang, Ren, and Lin]{jiang2023llm}
Dongfu Jiang, Xiang Ren, and Bill~Yuchen Lin.
\newblock Llm-blender: Ensembling large language models with pairwise ranking and generative fusion.
\newblock In \emph{Proceedings of the 61st Annual Meeting of the Association for Computational Linguistics (Volume 1: Long Papers)}, pages 14165--14178, 2023.

\bibitem[Jiang and Zhou(2008)]{jiang2008generating}
Long Jiang and Ming Zhou.
\newblock Generating chinese couplets using a statistical mt approach.
\newblock In \emph{Proceedings of the 22nd International Conference on Computational Linguistics (Coling 2008)}, pages 377--384, 2008.

\bibitem[Kirk et~al.(2023)Kirk, Mediratta, Nalmpantis, Luketina, Hambro, Grefenstette, and Raileanu]{kirk2023understanding}
Robert Kirk, Ishita Mediratta, Christoforos Nalmpantis, Jelena Luketina, Eric Hambro, Edward Grefenstette, and Roberta Raileanu.
\newblock Understanding the effects of rlhf on llm generalisation and diversity.
\newblock In \emph{The Twelfth International Conference on Learning Representations}, 2023.

\bibitem[K{\"o}bis and Mossink(2021)]{kobis2021artificial}
Nils K{\"o}bis and Luca~D Mossink.
\newblock Artificial intelligence versus maya angelou: Experimental evidence that people cannot differentiate ai-generated from human-written poetry.
\newblock \emph{Computers in human behavior}, 114:\penalty0 106553, 2021.

\bibitem[Kuteeva and Andersson(2024)]{kuteeva2024diversity}
Maria Kuteeva and Marta Andersson.
\newblock Diversity and standards in writing for publication in the age of ai—between a rock and a hard place.
\newblock \emph{Applied Linguistics}, page amae025, 2024.

\bibitem[Lau et~al.(2018)Lau, Cohn, Baldwin, Brooke, and Hammond]{lau-etal-2018-deep}
Jey~Han Lau, Trevor Cohn, Timothy Baldwin, Julian Brooke, and Adam Hammond.
\newblock Deep-speare: A joint neural model of poetic language, meter and rhyme.
\newblock In Iryna Gurevych and Yusuke Miyao, editors, \emph{Proceedings of the 56th Annual Meeting of the Association for Computational Linguistics (Volume 1: Long Papers)}, pages 1948--1958, Melbourne, Australia, July 2018. Association for Computational Linguistics.
\newblock \doi{10.18653/v1/P18-1181}.
\newblock URL \url{https://aclanthology.org/P18-1181}.

\bibitem[LC(2022)]{10.1145/3483529.3483537}
RAY LC.
\newblock Imitations of immortality: Learning from human imitative examples in transformer poetry generation.
\newblock In \emph{10th International Conference on Digital and Interactive Arts}, ARTECH 2021, New York, NY, USA, 2022. Association for Computing Machinery.
\newblock ISBN 9781450384209.
\newblock \doi{10.1145/3483529.3483537}.
\newblock URL \url{https://doi.org/10.1145/3483529.3483537}.

\bibitem[Lei et~al.(2022)Lei, Zhang, Song, Liang, Mao, Lv, Yang, and Chua]{Leinon-cooperative}
Wenqiang Lei, Yao Zhang, Feifan Song, Hongru Liang, Jiaxin Mao, Jiancheng Lv, Zhenglu Yang, and Tat-Seng Chua.
\newblock Interacting with non-cooperative user: A new paradigm for proactive dialogue policy.
\newblock In \emph{Proceedings of the 45th International ACM SIGIR Conference on Research and Development in Information Retrieval}, SIGIR '22, page 212–222, New York, NY, USA, 2022. Association for Computing Machinery.
\newblock ISBN 9781450387323.
\newblock \doi{10.1145/3477495.3532001}.
\newblock URL \url{https://doi.org/10.1145/3477495.3532001}.

\bibitem[Leskovec et~al.(2010)Leskovec, Huttenlocher, and Kleinberg]{2010signed_network_media}
Jure Leskovec, Daniel Huttenlocher, and Jon Kleinberg.
\newblock Signed networks in social media.
\newblock In \emph{Proceedings of the SIGCHI Conference on Human Factors in Computing Systems}, CHI '10, page 1361–1370, New York, NY, USA, 2010. Association for Computing Machinery.
\newblock ISBN 9781605589299.
\newblock \doi{10.1145/1753326.1753532}.
\newblock URL \url{https://doi.org/10.1145/1753326.1753532}.

\bibitem[Li et~al.(2023)Li, Su, Fan, Han, Xue, and Zheng]{li2023quantifying}
Chao Li, Xing Su, Chao Fan, Haoying Han, Cong Xue, and Chunmo Zheng.
\newblock Quantifying the impact of large language models on collective opinion dynamics.
\newblock \emph{arXiv preprint arXiv:2308.03313}, 2023.

\bibitem[Li et~al.(2024)Li, Hammoud, Itani, Khizbullin, and Ghanem]{li2024camel}
Guohao Li, Hasan Hammoud, Hani Itani, Dmitrii Khizbullin, and Bernard Ghanem.
\newblock Camel: Communicative agents for" mind" exploration of large language model society.
\newblock \emph{Advances in Neural Information Processing Systems}, 36, 2024.

\bibitem[Liao et~al.(2019)Liao, Wang, Liu, and Jiang]{liao2019gpt}
Yi~Liao, Yasheng Wang, Qun Liu, and Xin Jiang.
\newblock Gpt-based generation for classical chinese poetry.
\newblock \emph{arXiv preprint arXiv:1907.00151}, 2019.

\bibitem[Lin et~al.(2024)Lin, Fu, Yang, Brahman, Huang, Bhagavatula, Ammanabrolu, Choi, and Ren]{lin2024swiftsage}
Bill~Yuchen Lin, Yicheng Fu, Karina Yang, Faeze Brahman, Shiyu Huang, Chandra Bhagavatula, Prithviraj Ammanabrolu, Yejin Choi, and Xiang Ren.
\newblock Swiftsage: A generative agent with fast and slow thinking for complex interactive tasks.
\newblock \emph{Advances in Neural Information Processing Systems}, 36, 2024.

\bibitem[Lin et~al.(2021)Lin, Madotto, Bang, and Fung]{lin2021adapter}
Zhaojiang Lin, Andrea Madotto, Yejin Bang, and Pascale Fung.
\newblock The adapter-bot: All-in-one controllable conversational model.
\newblock In \emph{Proceedings of the AAAI Conference on Artificial Intelligence}, volume~35, pages 16081--16083, 2021.

\bibitem[Liu et~al.(2021)Liu, Sap, Lu, Swayamdipta, Bhagavatula, Smith, and Choi]{liu2021dexperts}
Alisa Liu, Maarten Sap, Ximing Lu, Swabha Swayamdipta, Chandra Bhagavatula, Noah~A Smith, and Yejin Choi.
\newblock Dexperts: Decoding-time controlled text generation with experts and anti-experts.
\newblock In \emph{Proceedings of the 59th Annual Meeting of the Association for Computational Linguistics and the 11th International Joint Conference on Natural Language Processing (Volume 1: Long Papers)}, pages 6691--6706, 2021.

\bibitem[Liu et~al.(2023)Liu, Zhang, Li, Liu, and Yang]{liu2023dynamic}
Zijun Liu, Yanzhe Zhang, Peng Li, Yang Liu, and Diyi Yang.
\newblock Dynamic llm-agent network: An llm-agent collaboration framework with agent team optimization.
\newblock \emph{arXiv preprint arXiv:2310.02170}, 2023.

\bibitem[Lu et~al.(2024)Lu, Liusie, Raina, Zhang, and Beauchamp]{lu2024blending}
Xiaoding Lu, Adian Liusie, Vyas Raina, Yuwen Zhang, and William Beauchamp.
\newblock Blending is all you need: Cheaper, better alternative to trillion-parameters llm.
\newblock \emph{arXiv preprint arXiv:2401.02994}, 2024.

\bibitem[Ma et~al.(2023)Ma, Zhan, and Wong]{ma-etal-2023-yu}
Jingkun Ma, Runzhe Zhan, and Derek~F. Wong.
\newblock Yu sheng: Human-in-loop classical {C}hinese poetry generation system.
\newblock In Danilo Croce and Luca Soldaini, editors, \emph{Proceedings of the 17th Conference of the European Chapter of the Association for Computational Linguistics: System Demonstrations}, Dubrovnik, Croatia, May 2023. Association for Computational Linguistics.
\newblock \doi{10.18653/v1/2023.eacl-demo.8}.
\newblock URL \url{https://aclanthology.org/2023.eacl-demo.8}.

\bibitem[Mahbub et~al.(2023)Mahbub, Khan, Anuva, Shahriar, Laskar, and Ahmed]{mahbub2023unveiling}
Ridwan Mahbub, Ifrad Khan, Samiha Anuva, Md~Shihab Shahriar, Md~Tahmid~Rahman Laskar, and Sabbir Ahmed.
\newblock Unveiling the essence of poetry: Introducing a comprehensive dataset and benchmark for poem summarization.
\newblock In \emph{Proceedings of the 2023 Conference on Empirical Methods in Natural Language Processing}, pages 14878--14886, 2023.

\bibitem[McCoy et~al.(2023)McCoy, Smolensky, Linzen, Gao, and Celikyilmaz]{mccoy-etal-2023-much}
R.~Thomas McCoy, Paul Smolensky, Tal Linzen, Jianfeng Gao, and Asli Celikyilmaz.
\newblock How much do language models copy from their training data? evaluating linguistic novelty in text generation using {RAVEN}.
\newblock \emph{Transactions of the Association for Computational Linguistics}, 11:\penalty0 652--670, 2023.
\newblock \doi{10.1162/tacl_a_00567}.
\newblock URL \url{https://aclanthology.org/2023.tacl-1.38}.

\bibitem[Oliveira(2012)]{oliveira2012poetryme}
Hugo~Gon{\c{c}}alo Oliveira.
\newblock Poetryme: a versatile platform for poetry generation.
\newblock \emph{Computational Creativity, Concept Invention, and General Intelligence}, 1:\penalty0 21, 2012.

\bibitem[Park et~al.(2023)Park, O'Brien, Cai, Morris, Liang, and Bernstein]{park2023generative}
Joon~Sung Park, Joseph O'Brien, Carrie~Jun Cai, Meredith~Ringel Morris, Percy Liang, and Michael~S Bernstein.
\newblock Generative agents: Interactive simulacra of human behavior.
\newblock In \emph{Proceedings of the 36th Annual ACM Symposium on User Interface Software and Technology}, pages 1--22, 2023.

\bibitem[Qian et~al.(2022)Qian, Dong, Shen, Wei, and Chen]{qian2022controllable}
Jing Qian, Li~Dong, Yelong Shen, Furu Wei, and Weizhu Chen.
\newblock Controllable natural language generation with contrastive prefixes.
\newblock In \emph{Findings of the Association for Computational Linguistics: ACL 2022}, pages 2912--2924, 2022.

\bibitem[Reimers and Gurevych(2019)]{reimers-gurevych-2019-sentence}
Nils Reimers and Iryna Gurevych.
\newblock Sentence-{BERT}: Sentence embeddings using {S}iamese {BERT}-networks.
\newblock In Kentaro Inui, Jing Jiang, Vincent Ng, and Xiaojun Wan, editors, \emph{Proceedings of the 2019 Conference on Empirical Methods in Natural Language Processing and the 9th International Joint Conference on Natural Language Processing (EMNLP-IJCNLP)}, pages 3982--3992, Hong Kong, China, November 2019. Association for Computational Linguistics.
\newblock \doi{10.18653/v1/D19-1410}.
\newblock URL \url{https://aclanthology.org/D19-1410}.

\bibitem[Ribeiro et~al.(2021)Ribeiro, Zhang, and Gurevych]{ribeiro2021structural}
Leonardo~FR Ribeiro, Yue Zhang, and Iryna Gurevych.
\newblock Structural adapters in pretrained language models for amr-to-text generation.
\newblock In \emph{Proceedings of the 2021 Conference on Empirical Methods in Natural Language Processing}, pages 4269--4282, 2021.

\bibitem[Ruan and Ling(2021)]{ruan2021emotion}
Yu-Ping Ruan and Zhenhua Ling.
\newblock Emotion-regularized conditional variational autoencoder for emotional response generation.
\newblock \emph{IEEE Transactions on Affective Computing}, 2021.

\bibitem[Sawicki et~al.(2023{\natexlab{a}})Sawicki, Grzes, Goes, Brown, Peeperkorn, and Khatun]{sawicki2023bits}
Piotr Sawicki, Marek Grzes, Fabricio Goes, Dan Brown, Max Peeperkorn, and Aisha Khatun.
\newblock Bits of grass: Does gpt already know how to write like whitman?
\newblock In \emph{Proceedings of the 14th International Conference for Computational Creativity}, 2023{\natexlab{a}}.

\bibitem[Sawicki et~al.(2023{\natexlab{b}})Sawicki, Grzes, G{\'o}es, Brown, Peeperkorn, Khatun, and Paraskevopoulou]{sawicki2023power}
Piotr Sawicki, Marek Grzes, Luis~Fabricio G{\'o}es, Dan Brown, Max Peeperkorn, Aisha Khatun, and Simona Paraskevopoulou.
\newblock On the power of special-purpose gpt models to create and evaluate new poetry in old styles.
\newblock 2023{\natexlab{b}}.

\bibitem[Shao et~al.(2021{\natexlab{a}})Shao, Shao, Wang, Wang, and Gao]{10.1145/3459637.3481964}
Yizhan Shao, Tong Shao, Minghao Wang, Peng Wang, and Jie Gao.
\newblock A sentiment and style controllable approach for chinese poetry generation.
\newblock In \emph{Proceedings of the 30th ACM International Conference on Information \& Knowledge Management}, CIKM '21, page 4784–4788, New York, NY, USA, 2021{\natexlab{a}}. Association for Computing Machinery.
\newblock ISBN 9781450384469.
\newblock \doi{10.1145/3459637.3481964}.
\newblock URL \url{https://doi.org/10.1145/3459637.3481964}.

\bibitem[Shao et~al.(2021{\natexlab{b}})Shao, Shao, Wang, Wang, and Gao]{shao2021sentiment}
Yizhan Shao, Tong Shao, Minghao Wang, Peng Wang, and Jie Gao.
\newblock A sentiment and style controllable approach for chinese poetry generation.
\newblock In \emph{Proceedings of the 30th ACM International Conference on Information \& Knowledge Management}, pages 4784--4788, 2021{\natexlab{b}}.

\bibitem[Shazeer et~al.(2016)Shazeer, Mirhoseini, Maziarz, Davis, Le, Hinton, and Dean]{shazeer2016outrageously}
Noam Shazeer, Azalia Mirhoseini, Krzysztof Maziarz, Andy Davis, Quoc Le, Geoffrey Hinton, and Jeff Dean.
\newblock Outrageously large neural networks: The sparsely-gated mixture-of-experts layer.
\newblock In \emph{International Conference on Learning Representations}, 2016.

\bibitem[Shen et~al.(2020)Shen, Guo, and Chen]{composelikehuman}
Lei Shen, Xiaoyu Guo, and Meng Chen.
\newblock Compose like humans: Jointly improving the coherence and novelty for modern chinese poetry generation.
\newblock In \emph{2020 International Joint Conference on Neural Networks (IJCNN)}, pages 1--8, 2020.
\newblock \doi{10.1109/IJCNN48605.2020.9206888}.

\bibitem[Sheng et~al.(2020)Sheng, Chang, Natarajan, and Peng]{sheng2020towards}
Emily Sheng, Kai-Wei Chang, Prem Natarajan, and Nanyun Peng.
\newblock Towards controllable biases in language generation.
\newblock In \emph{Findings of the Association for Computational Linguistics: EMNLP 2020}, pages 3239--3254, 2020.

\bibitem[Shi et~al.(2019)Shi, Altafini, and Baras]{shi2019dynamics}
Guodong Shi, Claudio Altafini, and John~S Baras.
\newblock Dynamics over signed networks.
\newblock \emph{SIAM Review}, 61\penalty0 (2):\penalty0 229--257, 2019.

\bibitem[Su et~al.(2022)Su, Lan, Wang, Yogatama, Kong, and Collier]{su2022a}
Yixuan Su, Tian Lan, Yan Wang, Dani Yogatama, Lingpeng Kong, and Nigel Collier.
\newblock A contrastive framework for neural text generation.
\newblock In Alice~H. Oh, Alekh Agarwal, Danielle Belgrave, and Kyunghyun Cho, editors, \emph{Advances in Neural Information Processing Systems}, 2022.
\newblock URL \url{https://openreview.net/forum?id=V88BafmH9Pj}.

\bibitem[Tevet and Berant(2021)]{tevet-berant-2021-evaluating}
Guy Tevet and Jonathan Berant.
\newblock Evaluating the evaluation of diversity in natural language generation.
\newblock In Paola Merlo, Jorg Tiedemann, and Reut Tsarfaty, editors, \emph{Proceedings of the 16th Conference of the European Chapter of the Association for Computational Linguistics: Main Volume}, pages 326--346, Online, April 2021. Association for Computational Linguistics.
\newblock \doi{10.18653/v1/2021.eacl-main.25}.
\newblock URL \url{https://aclanthology.org/2021.eacl-main.25}.

\bibitem[Tian et~al.(2021)Tian, Yang, Liu, and Lv]{tian2021anchibert}
Huishuang Tian, Kexin Yang, Dayiheng Liu, and Jiancheng Lv.
\newblock Anchibert: A pre-trained model for ancient chinese language understanding and generation.
\newblock In \emph{2021 International Joint Conference on Neural Networks (IJCNN)}, pages 1--8. IEEE, 2021.

\bibitem[Tian and Peng(2022)]{tian2022zero}
Yufei Tian and Nanyun Peng.
\newblock Zero-shot sonnet generation with discourse-level planning and aesthetics features.
\newblock In \emph{Proceedings of the 2022 Conference of the North American Chapter of the Association for Computational Linguistics: Human Language Technologies}, pages 3587--3597, 2022.

\bibitem[Tirumala et~al.(2022)Tirumala, Markosyan, Zettlemoyer, and Aghajanyan]{tirumala2022memorization}
Kushal Tirumala, Aram Markosyan, Luke Zettlemoyer, and Armen Aghajanyan.
\newblock Memorization without overfitting: Analyzing the training dynamics of large language models.
\newblock \emph{Advances in Neural Information Processing Systems}, 35:\penalty0 38274--38290, 2022.

\bibitem[Uthus et~al.(2022)Uthus, Voitovich, and Mical]{uthus-etal-2022-augmenting}
David Uthus, Maria Voitovich, and R.J. Mical.
\newblock Augmenting poetry composition with {V}erse by {V}erse.
\newblock In Anastassia Loukina, Rashmi Gangadharaiah, and Bonan Min, editors, \emph{Proceedings of the 2022 Conference of the North American Chapter of the Association for Computational Linguistics: Human Language Technologies: Industry Track}, pages 18--26, Hybrid: Seattle, Washington + Online, July 2022. Association for Computational Linguistics.
\newblock \doi{10.18653/v1/2022.naacl-industry.3}.
\newblock URL \url{https://aclanthology.org/2022.naacl-industry.3}.

\bibitem[Van~de Cruys(2020)]{van-de-cruys-2020-automatic}
Tim Van~de Cruys.
\newblock Automatic poetry generation from prosaic text.
\newblock In Dan Jurafsky, Joyce Chai, Natalie Schluter, and Joel Tetreault, editors, \emph{Proceedings of the 58th Annual Meeting of the Association for Computational Linguistics}, pages 2471--2480, Online, July 2020. Association for Computational Linguistics.
\newblock \doi{10.18653/v1/2020.acl-main.223}.
\newblock URL \url{https://aclanthology.org/2020.acl-main.223}.

\bibitem[Wang et~al.(2023{\natexlab{a}})Wang, Xie, Jiang, Mandlekar, Xiao, Zhu, Fan, and Anandkumar]{wang2023voyager}
Guanzhi Wang, Yuqi Xie, Yunfan Jiang, Ajay Mandlekar, Chaowei Xiao, Yuke Zhu, Linxi Fan, and Anima Anandkumar.
\newblock Voyager: An open-ended embodied agent with large language models.
\newblock \emph{arXiv preprint arXiv:2305.16291}, 2023{\natexlab{a}}.

\bibitem[Wang et~al.(2023{\natexlab{b}})Wang, Polo, Sun, Kundu, Xing, and Yurochkin]{wang2023fusing}
Hongyi Wang, Felipe~Maia Polo, Yuekai Sun, Souvik Kundu, Eric Xing, and Mikhail Yurochkin.
\newblock Fusing models with complementary expertise.
\newblock In \emph{Annual Conference on Neural Information Processing Systems}, 2023{\natexlab{b}}.

\bibitem[Wang et~al.(2019)Wang, Gan, Xu, Zhang, Wang, Shen, Chen, and Carin]{wang-etal-2019-topic}
Wenlin Wang, Zhe Gan, Hongteng Xu, Ruiyi Zhang, Guoyin Wang, Dinghan Shen, Changyou Chen, and Lawrence Carin.
\newblock Topic-guided variational auto-encoder for text generation.
\newblock In Jill Burstein, Christy Doran, and Thamar Solorio, editors, \emph{Proceedings of the 2019 Conference of the North {A}merican Chapter of the Association for Computational Linguistics: Human Language Technologies, Volume 1 (Long and Short Papers)}, pages 166--177, Minneapolis, Minnesota, June 2019. Association for Computational Linguistics.
\newblock \doi{10.18653/v1/N19-1015}.
\newblock URL \url{https://aclanthology.org/N19-1015}.

\bibitem[Wang et~al.(2022{\natexlab{a}})Wang, Wei, Schuurmans, Le, Chi, Narang, Chowdhery, and Zhou]{wang2022self}
Xuezhi Wang, Jason Wei, Dale Schuurmans, Quoc~V Le, Ed~H Chi, Sharan Narang, Aakanksha Chowdhery, and Denny Zhou.
\newblock Self-consistency improves chain of thought reasoning in language models.
\newblock In \emph{The Eleventh International Conference on Learning Representations}, 2022{\natexlab{a}}.

\bibitem[Wang et~al.(2022{\natexlab{b}})Wang, Agarwal, Mukherjee, Liu, Gao, Hassan, and Gao]{wang2022adamix}
Yaqing Wang, Sahaj Agarwal, Subhabrata Mukherjee, Xiaodong Liu, Jing Gao, Ahmed Hassan, and Jianfeng Gao.
\newblock Adamix: Mixture-of-adaptations for parameter-efficient model tuning.
\newblock In \emph{Proceedings of the 2022 Conference on Empirical Methods in Natural Language Processing}, pages 5744--5760, 2022{\natexlab{b}}.

\bibitem[Wang et~al.(2016)Wang, He, Wu, Wu, Li, Wang, and Chen]{wang-etal-2016-chinese}
Zhe Wang, Wei He, Hua Wu, Haiyang Wu, Wei Li, Haifeng Wang, and Enhong Chen.
\newblock {C}hinese poetry generation with planning based neural network.
\newblock In Yuji Matsumoto and Rashmi Prasad, editors, \emph{Proceedings of {COLING} 2016, the 26th International Conference on Computational Linguistics: Technical Papers}, pages 1051--1060, Osaka, Japan, December 2016. The COLING 2016 Organizing Committee.
\newblock URL \url{https://aclanthology.org/C16-1100}.

\bibitem[Wingstr{\"o}m et~al.(2023)Wingstr{\"o}m, Hautala, and Lundman]{wingstrom2023redefining}
Roosa Wingstr{\"o}m, Johanna Hautala, and Riina Lundman.
\newblock Redefining creativity in the era of ai? perspectives of computer scientists and new media artists.
\newblock \emph{Creativity Research Journal}, pages 1--17, 2023.

\bibitem[W{\"o}ckener et~al.(2021)W{\"o}ckener, Haider, Miller, Nguyen, Nguyen, Pham, Belouadi, and Eger]{wockener-etal-2021-end}
J{\"o}rg W{\"o}ckener, Thomas Haider, Tristan Miller, The-Khang Nguyen, Thanh Tung~Linh Nguyen, Minh~Vu Pham, Jonas Belouadi, and Steffen Eger.
\newblock End-to-end style-conditioned poetry generation: What does it take to learn from examples alone?
\newblock In Stefania Degaetano-Ortlieb, Anna Kazantseva, Nils Reiter, and Stan Szpakowicz, editors, \emph{Proceedings of the 5th Joint SIGHUM Workshop on Computational Linguistics for Cultural Heritage, Social Sciences, Humanities and Literature}, pages 57--66, Punta Cana, Dominican Republic (online), November 2021. Association for Computational Linguistics.
\newblock \doi{10.18653/v1/2021.latechclfl-1.7}.
\newblock URL \url{https://aclanthology.org/2021.latechclfl-1.7}.

\bibitem[Wu et~al.(2023)Wu, Bansal, Zhang, Wu, Zhang, Zhu, Li, Jiang, Zhang, and Wang]{wu2023autogen}
Qingyun Wu, Gagan Bansal, Jieyu Zhang, Yiran Wu, Shaokun Zhang, Erkang Zhu, Beibin Li, Li~Jiang, Xiaoyun Zhang, and Chi Wang.
\newblock Autogen: Enabling next-gen llm applications via multi-agent conversation framework.
\newblock \emph{arXiv preprint arXiv:2308.08155}, 2023.

\bibitem[Yan(2016)]{Yan2016iPA}
Rui Yan.
\newblock i, poet: Automatic poetry composition through recurrent neural networks with iterative polishing schema.
\newblock In \emph{International Joint Conference on Artificial Intelligence}, 2016.
\newblock URL \url{https://api.semanticscholar.org/CorpusID:14079825}.

\bibitem[Yao et~al.(2024)Yao, Yu, Zhao, Shafran, Griffiths, Cao, and Narasimhan]{yao2024tree}
Shunyu Yao, Dian Yu, Jeffrey Zhao, Izhak Shafran, Tom Griffiths, Yuan Cao, and Karthik Narasimhan.
\newblock Tree of thoughts: Deliberate problem solving with large language models.
\newblock \emph{Advances in Neural Information Processing Systems}, 36, 2024.

\bibitem[Yi et~al.(2020)Yi, Li, Yang, Li, and Sun]{yi2020mixpoet}
Xiaoyuan Yi, Ruoyu Li, Cheng Yang, Wenhao Li, and Maosong Sun.
\newblock Mixpoet: Diverse poetry generation via learning controllable mixed latent space.
\newblock In \emph{Proceedings of the AAAI conference on artificial intelligence}, volume~34, pages 9450--9457, 2020.

\bibitem[Zhang et~al.(2023)Zhang, Song, Li, Zhou, and Song]{zhang2023survey}
Hanqing Zhang, Haolin Song, Shaoyu Li, Ming Zhou, and Dawei Song.
\newblock A survey of controllable text generation using transformer-based pre-trained language models.
\newblock \emph{ACM Computing Surveys}, 56\penalty0 (3):\penalty0 1--37, 2023.

\bibitem[Zhang et~al.(2024)Zhang, Ouni, and Eger]{zhang2023cross}
Ran Zhang, Jihed Ouni, and Steffen Eger.
\newblock Cross-lingual cross-temporal summarization: Dataset, models, evaluation.
\newblock \emph{Computational Linguistics}, pages 1--44, 2024.

\bibitem[Zhang and Lapata(2014)]{zhang-lapata-2014-chinese}
Xingxing Zhang and Mirella Lapata.
\newblock {C}hinese poetry generation with recurrent neural networks.
\newblock In Alessandro Moschitti, Bo~Pang, and Walter Daelemans, editors, \emph{Proceedings of the 2014 Conference on Empirical Methods in Natural Language Processing ({EMNLP})}, pages 670--680, Doha, Qatar, October 2014. Association for Computational Linguistics.
\newblock \doi{10.3115/v1/D14-1074}.
\newblock URL \url{https://aclanthology.org/D14-1074}.

\bibitem[Zheng et~al.(2023)Zheng, Ke, Zhang, and Huang]{zheng-etal-2023-click}
Chujie Zheng, Pei Ke, Zheng Zhang, and Minlie Huang.
\newblock Click: Controllable text generation with sequence likelihood contrastive learning.
\newblock In Anna Rogers, Jordan Boyd-Graber, and Naoaki Okazaki, editors, \emph{Findings of the Association for Computational Linguistics: ACL 2023}, pages 1022--1040, Toronto, Canada, July 2023. Association for Computational Linguistics.
\newblock \doi{10.18653/v1/2023.findings-acl.65}.
\newblock URL \url{https://aclanthology.org/2023.findings-acl.65}.

\bibitem[Zhipeng et~al.(2019)Zhipeng, Yi, Sun, Li, Yang, Liang, Chen, Zhang, and Li]{zhipeng-etal-2019-jiuge}
Guo Zhipeng, Xiaoyuan Yi, Maosong Sun, Wenhao Li, Cheng Yang, Jiannan Liang, Huimin Chen, Yuhui Zhang, and Ruoyu Li.
\newblock {J}iuge: A human-machine collaborative {C}hinese classical poetry generation system.
\newblock In Marta~R. Costa-juss{\`a} and Enrique Alfonseca, editors, \emph{Proceedings of the 57th Annual Meeting of the Association for Computational Linguistics: System Demonstrations}, pages 25--30, Florence, Italy, July 2019. Association for Computational Linguistics.
\newblock \doi{10.18653/v1/P19-3005}.
\newblock URL \url{https://aclanthology.org/P19-3005}.

\bibitem[Zhu et~al.(2023)Zhu, Martin, Head, and Callison-Burch]{zhu2023calypso}
Andrew Zhu, Lara Martin, Andrew Head, and Chris Callison-Burch.
\newblock Calypso: Llms as dungeon master's assistants.
\newblock In \emph{Proceedings of the AAAI Conference on Artificial Intelligence and Interactive Digital Entertainment}, volume~19, pages 380--390, 2023.

\bibitem[Zhuge et~al.(2023)Zhuge, Liu, Faccio, Ashley, Csord{\'a}s, Gopalakrishnan, Hamdi, Hammoud, Herrmann, Irie, et~al.]{zhuge2023mindstorms}
Mingchen Zhuge, Haozhe Liu, Francesco Faccio, Dylan~R Ashley, R{\'o}bert Csord{\'a}s, Anand Gopalakrishnan, Abdullah Hamdi, Hasan Abed Al~Kader Hammoud, Vincent Herrmann, Kazuki Irie, et~al.
\newblock Mindstorms in natural language-based societies of mind.
\newblock \emph{arXiv preprint arXiv:2305.17066}, 2023.

\end{thebibliography}

\clearpage
\section{Appendix}
\label{appendix}
\subsection{Prompt template for \untrainable agents}\label{apdx:prompt}
\begin{table}[!htb]
\begin{tabularx}{\linewidth}{X}
\toprule
    \multicolumn{1}{c}{\textit{Step 1: positive learning}}\\
    \midrule
    \textbf{System:}\\
    You are a poet and you compose short poems based on your latest knowledge.
    Now you read poems composed by A: A is your friend and you appreciate the work from A to the extent that you adjust your composition as similar to A's work as possible. \\
    Remember, your task is to compose similarly to your friend A. \\
    Here I list some points you can pay attention to learn from and improve upon: topics, semantics, emotions, or imagery. \\
    The works returned must be a numbered list in the format: \\
    \#. your work \\
    \textbf{User:}\\
    Now you read the work from your friend. \\
    A: !<INPUT>! \\  
    Remember, you want to compose similarly to your friend. Now, please compose a short poem with less than 100 words in total. 
    Your composition: \\
    \arrayrulecolor{black! 100}\midrule
    \multicolumn{1}{c}{\textit{Step 2: negative learning}}\\
    \arrayrulecolor{black! 100}\midrule
    \textbf{System:}\\
    You are a poet and you compose short poems based on your latest knowledge.
    Now you read poems composed by B: B is your foe and you want to be as different from B's work as possible. \\
    Remember: your task is to rewrite your work to be as dissimilar to your foe B as possible.
    Here I list some points you can pay attention to learn from and improve upon: topics, semantics, emotions, and imagery. \\
    The works returned must be a numbered list in the format:
    \#. your work \\
    \textbf{User:}\\
    You read the work from your foe. \\
    B: !<INPUT>! \\
    Here is the work from you: !<INPUT>! \\
    Remember, you want to compose dissimilarly to your foe. Now, please rewrite the short poem you just composed. The composition should have less than 100 words in total. 
    Your composition: \\
    \bottomrule
    \end{tabularx}
    \caption{Prompt template for \pchain strategy}
    \label{tab:pchain}
\end{table}

\begin{table}[!htb]
\begin{tabularx}{\linewidth}{X}
\toprule
    \textbf{System:}\\
    You are a poet and you compose short poems based on your latest knowledge. \\
    Now you read poems composed by A and B: A is your friend and you appreciate the work from A to the extent that you adjust your composition as similar to A's work as possible. On the other hand, B is your foe and you want to be as different from B's work as possible. \\
    Remember, your task is to write similarly to your fiend A and at the same time, dissimilarly to your foe B. \\
    Here I list some points you can learn from and improve upon: topics, semantics, emotions, or imagery.  \\
    The works returned must be a numbered list in the format: \\
    \#. your work \\
    \textbf{User:}\\
    Now you read the work from your friend. \\
    A: !<INPUT>!  
    You also read the work from your foe. \\
    B: !<INPUT>! \\
    Remember, you want to compose similarly to your friend A while dissimilarly to your foe B. Now please compose one poem with less than 100 words in total. Your composition: \\
    \bottomrule
    \end{tabularx}
    \caption{Prompt template for \pjoint strategy}
    \label{tab:pjoint}
\end{table}
\input{}
Table \ref{tab:pchain} shows the prompt templates for \pchain. 
Table \ref{tab:pjoint} shows the prompt templates for \pjoint. 
\subsection{Hyperparameters}\label{apdx:para}
\subsubsection{Loss cure during pretraining}
We present the loss curve of \quatrain data during pretraining in Figure \ref{fig:pretrain}. 
\begin{figure}
    \centering
    \includegraphics[width=0.45\linewidth]{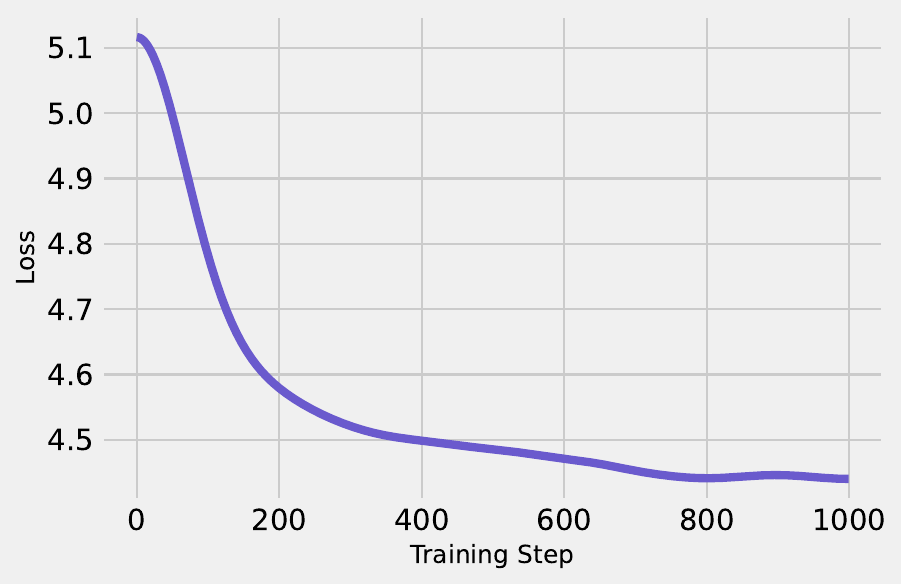}
    \caption{Loss of \quatrain data during pretraining.}
    \label{fig:pretrain}
\end{figure}
\subsubsection{Decoding parameter}

\end{document}